\documentclass{article}
\usepackage[preprint]{Arxiv_2026}

\usepackage[utf8]{inputenc}
\usepackage[T1]{fontenc}
\usepackage{microtype}

\usepackage{amsmath,amssymb,amsfonts,mathtools,bm}
\usepackage{nicefrac}

\usepackage{booktabs}
\usepackage{array}
\usepackage{longtable}
\usepackage{multirow}
\usepackage{enumitem}
\usepackage{float}
\usepackage{caption}

\usepackage{algorithm}
\usepackage{algpseudocode}

\usepackage[dvipsnames]{xcolor}
\usepackage{tikz}
\usepackage{pgfplots}
\usepgfplotslibrary{groupplots}
\usetikzlibrary{
  arrows.meta, positioning, shapes.geometric, shapes.multipart,
  fit, backgrounds, calc, decorations.pathreplacing,
  decorations.markings, chains, matrix, decorations.pathmorphing
}

\usepackage{hyperref}
\usepackage{url}
\hypersetup{colorlinks=true, linkcolor=blue, citecolor=blue, urlcolor=blue}
\definecolor{mambaBlue}{RGB}{52,120,190}
\definecolor{flowOrange}{RGB}{210,95,45}
\definecolor{physGreen}{RGB}{60,150,80}
\definecolor{gruPurple}{RGB}{160,60,170}
\definecolor{lossRed}{RGB}{200,50,50}
\definecolor{boxBg}{RGB}{248,249,252}

\tikzset{
  gbox/.style={rectangle, draw, rounded corners=3pt,
    minimum width=2.6cm, minimum height=0.65cm,
    align=center, font=\small, inner sep=4pt},
  encBox/.style={gbox, fill=mambaBlue!12,  draw=mambaBlue!70!black},
  physBox/.style={gbox, fill=physGreen!12, draw=physGreen!70!black},
  flowBox/.style={gbox, fill=flowOrange!12,draw=flowOrange!70!black},
  gruBox/.style={gbox, fill=gruPurple!12,  draw=gruPurple!70!black},
  lossBox/.style={gbox, fill=lossRed!10,   draw=lossRed!60},
  ioBox/.style={gbox, fill=gray!10,        draw=gray!50},
  arr/.style ={-{Stealth[length=2.5mm, width=1.8mm]}, thick},
  darr/.style={-{Stealth[length=2.5mm, width=1.8mm]}, thick, dashed, gray!60},
  plainbox/.style={draw, rounded corners, align=center,
    minimum width=2.45cm, minimum height=9mm}
}

\usepackage{amsmath,amssymb}
\usepackage{xcolor}
\usetikzlibrary{arrows.meta,decorations.pathreplacing,calc}

\definecolor{ink}{RGB}{35,39,47}
\definecolor{muted}{RGB}{100,110,120}
\definecolor{baseblue}{RGB}{92,145,210}
\definecolor{targetrose}{RGB}{190,80,105}
\definecolor{flowteal}{RGB}{35,135,125}
\definecolor{carbamber}{RGB}{200,110,30}
\definecolor{insblue}{RGB}{45,100,180}
\definecolor{penaltyred}{RGB}{220,50,50}

\newcommand{\Ljac}{\mathcal{L}_{\mathrm{jac}}}

\newcommand{\doop}{\operatorname{do}}

\title{Interventional Flow Matching: Prospective Dose-Response Forecasting with Velocity-Field Jacobian Regularization}

\author{%
Amirreza Dolatpour Fathkouhi\\
Department of Computer Science\\
Center for Diabetes Technology\\
University of Virginia\\
Charlottesville, VA, USA\\
\texttt{aww9gh@virginia.edu}
\And
Justin Lee\\
School of Data Science\\
Center for Diabetes Technology\\
University of Virginia\\
Charlottesville, VA, USA\\
\texttt{jgh2xh@virginia.edu}
\AND
Heman Shakeri\\
School of Data Science\\
Center for Diabetes Technology\\
University of Virginia\\
Charlottesville, VA, USA\\
\texttt{hs9hd@virginia.edu}
}

\begin{document}
\maketitle

\begin{abstract}
Predicting a patient's physiological trajectory under a planned treatment sequence is a prospective interventional problem, not standard time-series extrapolation. We study this problem in glucose management, where insulin and carbohydrate records are policy-dependent: future drivers are coupled to patient state, behavior, and clinical decision rules, so observational forecasting accuracy alone does not guarantee correct responses to planned interventions.

We introduce Interventional Flow Matching (IFM), a continuous-time generative framework for physiologically constrained prospective forecasting. IFM conditions a flow-matching velocity field on patient history and planned future drivers in a bounded latent glucose space. Rather than embedding strict mechanistic glucose--insulin ODE equations or enforcing causality through rollout-based simulations, IFM uses a solver-free regularization: it penalizes the Jacobian of the instantaneous velocity field with respect to smoothed treatment drivers. This imposes signed, dose-bounded local sensitivities directly on the learned dynamics: insulin lowers glucose, carbohydrates raise it, and both responses remain within plausible ranges.

On a simulated UVA/Padova type 1 diabetes cohort, IFM achieves the strongest balance between observed-driver RMSE and interventional response metrics. Across experiments, it consistently produces physiologically correct responses to both insulin and carbohydrate drivers while maintaining high directional, and ranking consistency.
\end{abstract}

\section{Introduction}
\label{sec:intro}
Glucose forecasting for treatment planning is a \emph{prospective interventional} problem. At decision time, future insulin and carbohydrate inputs are not observed; they are candidate actions. Given the observed history \(X_{1:L}\), containing past glucose, insulin, and carbohydrate measurements, the goal is to predict the glucose trajectory under a planned future driver sequence \(U_{1:H}^{\mathrm{int}}\).

In observational data, future drivers follow the clinical or behavioral policy \(U_{1:H}\sim \pi_{\mathrm{obs}}(\cdot\mid X_{1:L})\). At deployment, this natural assignment process is replaced by a chosen plan, \(U_{1:H}\gets U_{1:H}^{\mathrm{int}}\). Following Pearl's \(\doop\)-operator~\citep{pearl2009causality}, we aim to learn the target \(p(G_{1:H}\mid X_{1:L},\doop(U_{1:H}=U_{1:H}^{\mathrm{int}}))\). The model must therefore learn the \emph{physiological effect of imposed treatments}, not only the correlations present in observed treatment patterns. 

This is difficult because glucose data are strongly confounded. Insulin is often delivered when glucose is already rising, and meals occur in contexts that also influence future glucose. A purely predictive model can fit observed trajectories while learning incorrect intervention semantics, such as associating insulin with rising glucose even though its physiological effect is glucose-lowering.

Exact glucose forecasting under planned interventions is intrinsically difficult because future glucose is affected by many unobserved or poorly measured factors, including activity, stress, illness, absorption variability, and recording error. Therefore, the most defensible intervention claim in this setting is not exact interventional trajectory accuracy, but whether the model predicts plausible response magnitude, correct response direction, and consistent dose ranking. This also matches the intended use of interventional trajectory forecasting: comparing generated trajectories under alternative treatment plans and identifying those that move glucose toward a desired range~\citep{lee2024shortcomings}. For this reason, sensitivity, directional validity, and ranking validity are central requirements for prospective diabetes intervention forecasting, beyond trajectory accuracy alone.

We propose \emph{Interventional Flow Matching} (IFM), a conditional flow matching model~\citep{lipman2023flow} for prospective glucose response prediction. IFM represents glucose in an unbounded latent space and constrains the \emph{local flow response} to planned interventions. Future insulin and carbohydrate inputs are pharmacokinetically smoothed and injected through an AdaLN-modulated velocity pathway~\citep{zhang2025dim}. A velocity-field Jacobian penalty then enforces physiologically valid sensitivities: insulin should lower glucose, carbohydrates should raise it, and both effects should remain within plausible physiological sensitivity bounds. \emph{This uses weak directional physiological knowledge without requiring a full mechanistic ODE simulator.} This response penalty is \emph{simulation-free}: it is applied to the instantaneous velocity field at sampled flow-matching states, rather than through full alternative-intervention rollouts as in ranking-based objectives~\citep{zou2024hybrid}.

The full modeling framework performs a trajectory rollout by delegating i) autoregressive physical time advancement to a recurrent cell which updates history and context information that ii) gets passed into the IFM module to generate the glucose prediction at the current physical time. This prediction is then input to the recurrent cell for the next time step, and so on. We employ a small subsampled rollout loss only to control autoregressive error accumulation. Empirically, IFM maintains competitive observed-driver RMSE while improving
interventional metrics, with physiologically correct contributions from both
insulin and carbohydrate drivers.

\paragraph{Contributions.}
This paper makes four main contributions:

\begin{itemize}[leftmargin=*, itemsep=2pt, topsep=2pt]

  \item \textbf{Physiologically valid interventional flow responses.}
  We propose IFM, which routes planned insulin and carbohydrate inputs through the conditional flow-matching velocity field and regularizes its \emph{local response} to interventions. Instead of relying only on observational trajectory fitting, IFM constrains intervention sensitivities to have physiologically valid signs and magnitudes: insulin lowers glucose, carbohydrates raise it, and both responses remain within prescribed physiological sensitivity bounds.

  \item \textbf{Physiological response constraints without explicit mechanistic ODEs.}
  Unlike hybrid physiological ODE models, IFM does not require hand-specified glucose--insulin differential equations, latent physiological compartments, or graph-structured mechanistic interactions. It uses weak but clinically reliable directional knowledge about intervention effects, making it a lighter alternative to fully mechanistic physiological scaffolds.

  \item \textbf{Simulation-free intervention regularization.}
  Compared with ranking-based causal objectives that require rollout simulations to evaluate intervention effects, IFM applies a Jacobian regularizer directly to the instantaneous velocity field. This makes the physiological-response penalty solver-free, while a small subsampled fidelity term controls autoregressive rollout error.

\item \textbf{Improved accuracy--intervention tradeoff.}
On the UVA/Padova type 1 diabetes benchmark, IFM maintains high observed-driver
forecasting accuracy while producing physiologically correct responses to both
drivers, with insulin-lowering and carbohydrate-raising effects alongside high
directional and ranking consistency.

\end{itemize}
\section{Related Work}
\label{sec:related}

\paragraph{Longitudinal causal inference.}
Treatment-effect estimation under time-varying confounding uses propensity adjustment, neural G-computation, and adversarial balancing~\citep{bica2020estimating,li2021gnet,melnychuk2022causal,xia2023neural}. These methods mainly regularize history representations to address treatment-assignment bias. IFM is complementary: it leaves the encoder unconstrained and instead regularizes the velocity field that generates prospective forecasts.

\paragraph{Mechanistic and hybrid glucose models.}
Graph-based neural simulators use message passing to model physical dynamics~\citep{allen2023graph,wu2022learning,pfaff2021learning,li2022graph}. In glucose forecasting, hybrid causal models impose interventional structure through mechanistic ODEs, latent compartments, learnable physiological parameters, or rollout-based ranking losses~\citep{zou2024hybrid,zou2026automatic,miller2020learning}. These methods encode detailed physiological assumptions. IFM takes a lighter approach: it uses only directional priors---insulin lowers glucose and carbohydrates raise it---and enforces them with a local velocity-field Jacobian penalty rather than a full physiological scaffold; rollout simulation is used separately for forecasting and fidelity supervision, not to impose the physiological response constraint.

\paragraph{Flow matching and structured intervention regularization.}
Flow matching learns a velocity field that transports samples from a base distribution to the data distribution~\citep{lipman2023flow,tamir2024conditional,scassola2026graph}. In conditional forecasting, planned treatments enter this velocity field as conditioning variables, so derivatives of the velocity field with respect to
these treatment inputs describe the model's local interventional response. Prior causal regularizers shape model behavior by suppressing nuisance features or improving causation and dose--response estimation~\citep{dong2024background,bahadori2017causal,nie2021vcnet}. This motivates IFM's use of the instantaneous velocity field as the site of solver-free intervention regularization. Unlike DoFlow, which uses graph-conditioned invertible flows and factual-future abduction for counterfactual prediction~\citep{wu2025doflow}, IFM targets prospective forecasting without requiring a causal graph, and regularizes the velocity-field Jacobian to enforce signed physiological responses to smoothed insulin and carbohydrate drivers.

\section{Interventional Flow Matching}
\label{sec:method}

\begin{figure}[t]
    \centering
    \resizebox{\linewidth}{!}{\input{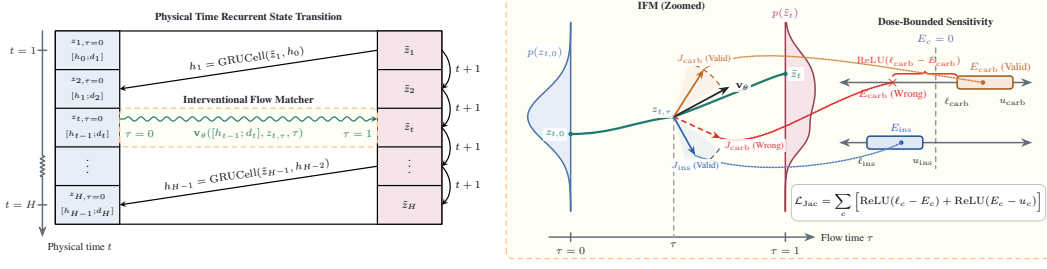}}
    \caption{
    Illustration of the full modeling framework. The $\mathrm{GRUCell}$ progresses physical time $t$ and the IFM module generates predictions through flow time $\tau$. IFM measures local insulin and carbohydrate sensitivities of the velocity field, converts them to physical glucose units, and penalizes values outside signed physiological sensitivity bounds: negative for insulin and positive for carbohydrates.
    }
    \label{fig:ifm_diagram}
\end{figure}

We instantiate the prospective interventional problem with continuous glucose $G_t$ as the response, and planned treatments $U_t=[u_t^{\mathrm{ins}},u_t^{\mathrm{carb}}]$ as a two-channel insulin--carbohydrate driver.
Given an observed patient history
$X_{1:L}\in\mathbb{R}^{L\times d_{\mathrm{in}}}$ with $d_{\mathrm{in}}=3$ denoting insulin, carbohydrate, and glucose history channels, and a planned future intervention sequence
$U_{1:H}^{\mathrm{int}}\in\mathbb{R}^{H\times d_u}$ with $d_u=2$ denoting insulin and carbohydrate drivers, IFM learns the interventional forecasting map
\[
    (X_{1:L},U_{1:H}^{\mathrm{int}})\mapsto \hat G_{1:H}^{\mathrm{int}} \, ,
\]
where $\hat G_{1:H}^{\mathrm{int}}\in\mathbb{R}^{H}$ is the predicted glucose trajectory under the specified treatment plan.
The goal is robust generalization to intervention plans not observed during training. 

\subsection{Input Preprocessing}
\paragraph{Mamba history encoder.}
\label{sec:encoder}

IFM uses a residual Mamba encoder~\citep{gu2023mamba} to summarize a batch of observed histories \(X_{1:L}\in\mathbb{R}^{B\times L\times d_{\mathrm{in}}}\). The encoder maps each input step to dimension \(d_m\), processes the sequence with RMSNorm-preconditioned~\citep{zhang2019root} Mamba blocks, and uses the final state \(h_0\in\mathbb{R}^{B\times d_m}\) to initialize the recurrent forecasting loop. Full architectural details are provided in Appendix~\ref{apx:encoder}.

\paragraph{PharmacoKinetic smoothing of future drivers (PK smoother).}
\label{sec:pksmooth}

Insulin and carbohydrate inputs arrive as sparse impulses, but their physiological effects are delayed and distributed over time. IFM therefore maps the planned driver sequence \(U_{1:H}^{\mathrm{int}}\in\mathbb{R}^{B\times H\times d_u}\) into an effective smoothed driver sequence
\(D_{1:H}=(d_1,\dots,d_H)\in\mathbb{R}^{B\times H\times d_u}\).
The smoother uses learned per-channel fast and slow decay states to produce a delayed-peak response, providing physiologically informed insulin and carbohydrate features for the velocity network. Full details are provided in Appendix~\ref{apx:pksmooth}.

\subsection{Unbounded Latent Glucose Representation}
\label{sec:latent}

Rather than modeling glucose directly in physical units, IFM maps glucose into an unconstrained latent space and decodes it through a bounded sigmoid transform. For a glucose target \(g\in[g_{\min},g_{\max}]\), we compute
\begin{equation}
    y =
    \left(
        \frac{g-g_{\min}}
        {g_{\max}-g_{\min}}
    \right)^{\kappa},
    \qquad
    z=\operatorname{logit}(y) \, ,
    \label{eq:encode}
\end{equation}
where \(\kappa>0\) is a fixed skew hyperparameter controlling how much of the latent range \(z\) maps to interior values of \(g\).
For a horizon target, this produces latent glucose \(z_{1:H}\in\mathbb{R}^{B\times H}\).
The inverse map
\begin{equation}
    g=g_{\min}+(g_{\max}-g_{\min})y^{1/\kappa}
    \label{eq:decode}
\end{equation}
decodes latent predictions back to physical glucose using \(y=\sigma(z)=(1+e^{-z})^{-1}\).
Thus, all flow dynamics operate in the unbounded latent space \(z\), while decoded predictions remain bounded within the physiological range \([g_{\min},g_{\max}]\).
For the remainder of this paper we fix \(\kappa = 1\), but additional ablations on the effect of \(\kappa\) are provided in Appendix~\ref{apx:skew_param}.

\subsection{Conditional Velocity Network}
\label{sec:velnet}

At each forecast step \(t \in [1,H]\), IFM concatenates the recurrent state \(h_{t-1}\in\mathbb{R}^{B\times d_m}\) with the smoothed driver \(d_t\in\mathbb{R}^{B\times d_u}\) to form the conditioning vector \(h_{t-1}^{\mathrm{cond}}=[h_{t-1};d_t]\in\mathbb{R}^{B\times d_{\mathrm{cond}}}\), where \(d_{\mathrm{cond}}=d_m+d_u\). This vector modulates a scalar velocity field \(v_\theta:\mathbb{R}^{d_{\mathrm{cond}}}\times\mathbb{R}\times[0,1]\to\mathbb{R}\), which transports latent glucose in continuous flow time.

Flow time \(\tau\in[0,1]\) is encoded with \(K\) Fourier features~\citep{tancik2020fourier}, denoted by \(\psi(\tau)\in\mathbb{R}^{B\times(1+2K)}\). Given a latent state \(z_{t,\tau}\in\mathbb{R}^{B}\) at times $(t, \tau)$, the network forms \([z_{t,\tau};\psi(\tau)]\in\mathbb{R}^{B\times(2+2K)}\) and computes

\begin{align}
    r
        &= \mathrm{Linear}_{2+2K\rightarrow d_f}
           \bigl([z_{t,\tau};\psi(\tau)]\bigr) \, ,
        & r\in\mathbb{R}^{B\times d_f} \, , \\
    [\gamma;\beta]
        &= \mathrm{Linear}_{d_{\mathrm{cond}}\rightarrow 2d_f}
           \bigl(h_{t-1}^{\mathrm{cond}}\bigr) \, ,
        & \gamma,\beta\in\mathbb{R}^{B\times d_f} \, , \\
    \tilde r
        &= \mathrm{LayerNorm}\!\left(\mathrm{SiLU}(r)\right)
           \odot (1+\gamma) + \beta \, ,
        & \tilde r\in\mathbb{R}^{B\times d_f} \, , \\
    v_\theta(h_{t-1}^{\mathrm{cond}},z_{t,\tau},\tau)
        &= \mathrm{Linear}_{d_f\rightarrow 1}
           \left(\mathrm{MLP}_{d_f\rightarrow d_f}(\tilde r)\right) \, ,
        & v_\theta(h_{t-1}^{\mathrm{cond}},z_{t,\tau},\tau)\in\mathbb{R}^{B} \, .
\end{align}
Here \(d_f\) is the hidden width of the velocity network, and \(\mathrm{MLP}_{d_f\rightarrow d_f}\) denotes the stack of hidden linear layers with SiLU activations. Since LayerNorm has no learnable affine parameters, the AdaLN~\cite{zhang2025dim} parameters \((\gamma,\beta)\) provide the context-dependent rescaling and shifting. Thus, changes in the planned driver \(d_t\) directly alter the local latent transport dynamics.

\subsection{Autoregressive Rollout and Teacher-Forced Conditioning}
\label{sec:rollout}

\paragraph{Driver-isolated recurrent update.}
After each prediction, the recurrent state is advanced using only latent glucose via
\begin{equation}
    h_{t}
    =
    \mathrm{GRUCell}\!\left(\tilde z_t,h_{t-1}\right) \, ,
    \label{eq:gru}
\end{equation}
where \(\tilde{z}_t \in \mathbb{R}^{B}\) is either the predicted latent glucose \(\hat{z}_t\) or the ground-truth latent glucose \(z_t^\star\) under teacher forcing.
Thus, planned interventions affect the forecast through just the AdaLN-modulated velocity field \(v_\theta(h_{t-1}^{\mathrm{cond}},z_{t, \tau},\tau)\), not the recurrent state transition.

\paragraph{Teacher-forced conditioning.}
During training, we use teacher-forced recurrent updates which provides two major benefits. First, the ground truth \(z_t^\star\) provides a stable conditioning trajectory. Second, access to the full \(z_t^\star\) trajectory allows us to bypass the integration step in generating a prediction \(\hat{z}_t\) for the recurrent state update,
\(
    h_{t}^{\mathrm{TF}}
    =
    \mathrm{GRUCell}(z_t^\star,h_{t-1}^{\mathrm{TF}})
\)
where we pair the previous teacher-forced state \(h_{t-1}^{\mathrm{TF}}\) with the smoothed driver to form
\(h_{t-1}^{\mathrm{TF,cond}}=[h_{t-1}^{\mathrm{TF}};d_{t}]\). Stacking over the horizon gives
\(h_{\mathrm{all}}^{\mathrm{TF,cond}}\in\mathbb{R}^{B\times H\times d_{\mathrm{cond}}}\), which conditions the flow matching and Jacobian objectives (Section~\ref{sec:jac}).

\paragraph{Autoregressive inference.}
At inference time, IFM generates forecasts autoregressively over physical time $t$.
Given the conditioning vector \(h_{t-1}^{\mathrm{cond}}\) and a base noise sample \(z_{t,0}\in\mathbb{R}^{B}\) from \(\mathcal{N}(0,1)\), IFM integrates the learned velocity field \(v_\theta\) with \(K_f\) forward-Euler steps starting at $k = 0$:
\begin{equation}
    z_{t,\tau_{k+1}} = z_{t,\tau_{k}}
        + \Delta\tau\,
        v_\theta\!\left(h_{t-1}^{\mathrm{cond}},z_{t,\tau_{k}},\tau_k\right),
    \qquad
    \tau_k=\frac{k}{K_f},
    \qquad
    \Delta\tau = \frac{1}{K_f}
    \label{eq:euler}
\end{equation}
and uses the resulting $\hat{z}_{t, 1}$ to update the recurrent state \( h_{t} = \mathrm{GRUCell}(\hat{z}_t, h_{t-1}) \). The full autoregressive interventional rollout is summarized in
Appendix~\ref{apx:inference}, Algorithm~\ref{alg:interventional}.

\section{Training Objectives}
\label{sec:training}

Training combines three complementary objectives for accurate, physiologically constrained, and rollout-consistent forecasting. The flow matching ($\mathcal{L}_{\mathrm{CFM}}$) and Jacobian (\(\mathcal{L}_{\mathrm{jac}}\)) objectives act locally on the instantaneous velocity field using the teacher-forced conditioning tensor \(h_{\mathrm{all}}^{\mathrm{TF,cond}}\) so they do not require full rollout integration during optimization. Subsequently, a lightweight simulation fidelity penalty ($\mathcal{L}_{\mathrm{fid}}$) supervises the multi-step autoregressive forecast on a subsampled batch, discouraging error accumulation during free-running inference. The total training loss is
\begin{equation}
    \mathcal{L}=\mathcal{L}_{\mathrm{CFM}}+\lambda_{\mathrm{jac}}\mathcal{L}_{\mathrm{jac}}+\lambda_{\mathrm{fid}}\mathcal{L}_{\mathrm{fid}}
\end{equation}
where \(\lambda_{\mathrm{jac}}\) and \(\lambda_{\mathrm{fid}}\) are scalar weights. We describe the individual components below.

\subsection{Conditional Flow Matching (CFM) Loss}
\label{sec:cfm}

The primary generative objective trains the velocity network to transport Gaussian base noise to the target latent glucose distribution, conditioned on patient history and the planned intervention. Following the linear-interpolant formulation of flow matching~\citep{lipman2023flow}, we encode the ground-truth future glucose into latent space to obtain \(z_{1:H}^\star\in\mathbb{R}^{B\times H}\). We then sample \(z_{1:H,0}\sim\mathcal{N}(0,1)\) and draw independent flow times
\(\tau^{(b,t)}\sim\mathrm{Uniform}(0,1)\) for each sample \(b\) and horizon step \(t\):

\[
    z_{t,\tau}^{(b)}
    =
    (1-\tau^{(b,t)})z_{t,0}^{(b)}
    +
    \tau^{(b,t)}z_t^{\star (b)},
    \qquad
    v_t^{\star\,(b)}=z_t^{\star(b)}-z_{t,0}^{(b)} \, .
\]

Using \(h_{\mathrm{all}}^{\mathrm{TF,cond}}\), the conditional flow matching loss is
\begin{equation}
    \mathcal{L}_{\mathrm{CFM}}
    =
    \frac{1}{BH}
    \sum_{b=1}^{B}\sum_{t=1}^{H}
    \left\|
        v_\theta\!\left(
            h_{\mathrm{all}}^{\mathrm{TF}, \mathrm{cond},(b,t)},
            z_{t, \tau}^{(b)},
            \tau^{(b, t)}
        \right)
        -
        v_{t}^{\star\,(b)}
    \right\|_2^2 \, .
    \label{eq:lcfm}
\end{equation}

\subsection{Velocity-Field Jacobian Regularization}
\label{sec:jac}
 
CFM models observational dynamics, but does not constrain
intervention semantics. Under policy confounding, a model can achieve
low CFM loss while ignoring treatment inputs entirely, or worse,
learning inverted responses (e.g., predicting glucose to rise after
insulin)~\citep{shakeri2025driver}. To impose a structural physiological
prior on this response, we regularize the gradient of the velocity field
with respect to the planned drivers and ask that each component of that
gradient lies within clinically motivated bounds.
 
\paragraph{Single-pass computation of pointwise driver Jacobians.}
To isolate the direct intervention effect, we stop gradients through the recurrent state \( h_{t-1}^{\mathrm{TF}} \),
the interpolated latent glucose \( z_{t, \tau}^{(b)} \), and the flow time \( \tau \).
We then aggregate the velocity outputs into a single scalar
\begin{equation}
    S
    =
    \sum_{b=1}^{B}\sum_{t=1}^{H}
    v_\theta\!\left(
        [h_{t-1}^{\mathrm{TF,sg},(b)};\,d_t^{(b)}],\,
        z_{t,\tau}^{\mathrm{sg},(b)},\,
        \tau^{\mathrm{sg},(b,t)}
    \right) \, ,
    \label{eq:jacobian_scalar}
\end{equation}
where $(\cdot)^{\mathrm{sg}}$ denotes a stopped gradient, and
differentiate once with respect to the entire driver tensor:
\(
    J = \frac{\partial S}{\partial D_{1:H}}
    \in\mathbb{R}^{B\times H\times d_u} \, .
\)
Although we compute these derivatives by differentiating a summed
scalar, separability of the velocity evaluations across $(b,t)$ means
the resulting tensor contains the \emph{pointwise local} derivatives
\begin{equation}
    J_c^{(b,t)}
    =
    \frac{\partial S}{\partial d_t^{(b,c)}}
    =
    \frac{\partial v_\theta\!\bigl(
        [h_{t-1}^{\mathrm{TF,sg},(b)};\,d_t^{(b)}],\,
        z_{t,\tau}^{\mathrm{sg},(b)},\,
        \tau^{\mathrm{sg},(b,t)}\bigr)}
    {\partial d_t^{(b,c)}} \, ,
    \qquad
    c\in\{\mathrm{ins},\mathrm{carb}\} \, .
\end{equation}
The summation in Eq.~\eqref{eq:jacobian_scalar} is an autograd batching
device, not a cumulative-effect operator; each $J_c^{(b,t)}$ depends
only on the velocity evaluation at $(b,t)$. The penalty acts on the
smoothed effective driver $d_t$ rather than the raw impulse $u_t$;
gradients through the PK smoother are stopped at this stage so the
smoother is not updated by the Jacobian regularizer, although it
remains learnable through the CFM and fidelity objectives.

\paragraph{Local sensitivity response constraints.}
The velocity field operates in latent \(z\)-space, whereas physiological
constraints are specified in mg/dL. We convert each local driver derivative \(J_c^{(b, t)}\) to physical glucose units using the analytic decoder slope derived from the latent transform in
Eq.~\eqref{eq:encode}:
\begin{equation}
    \frac{dg}{dz}
    =
    \frac{g_{\max}-g_{\min}}{\kappa}
    y^{1/\kappa}(1-y),
    \label{eq:slope}
\end{equation}
\begin{equation}
    E_c^{(b,t)}
    =
    \alpha_c\,J_c^{(b,t)}
    \frac{dg}{dz}\!\left(z_{t,\tau}^{(b)}\right) \, .
    \label{eq:sensitivity}
\end{equation}
Here \(\alpha_{\mathrm{ins}}=1.0\) unit of insulin and \(\alpha_{\mathrm{carb}}=25.0\) grams of carbohydrate map the insulin and carbohydrate perturbation units onto a common mg/dL response scale. We impose physiologically motivated sensitivity bounds \([\ell_c,u_c]\) on both channels: insulin sensitivity must be negative, \(\ell_{\mathrm{ins}}<u_{\mathrm{ins}}<0\), while carbohydrate sensitivity must be positive, \(0<\ell_{\mathrm{carb}}<u_{\mathrm{carb}}\). Because both valid intervals exclude zero, the regularizer also penalizes the lazy solution \(E_c\approx0\), where the model ignores either intervention channel instead of learning a physiologically meaningful response. We interpret these as \emph{local effective-driver sensitivity responses} evaluated at sampled CFM flow states, not as endpoint cumulative dose-response bounds. The regularizer applies a two-sided hinge per component of \(E\):

\begin{equation}
    \Ljac
    =
    \frac{1}{BH}
    \sum_{b=1}^{B}\sum_{t=1}^{H}\sum_{c}
    \left[
        \mathrm{ReLU}\!\left(\ell_c - E_c^{(b,t)}\right)
        +
        \mathrm{ReLU}\!\left(E_c^{(b,t)} - u_c\right)
    \right] \, .
    \label{eq:ljac}
\end{equation}
Because the hinge acts pointwise on each $E_c^{(b,t)}$, a wrong-signed
local response (e.g., a brief upward velocity response to insulin) is
penalized at the step where it occurs. A cumulative-aggregate hinge on
$\sum_t E_c^{(b,t)}$ would, by contrast, allow such a wrong-signed
contribution to be masked by larger correctly-signed contributions
elsewhere; the pointwise version removes that cancellation
channel.

\paragraph{What the penalty does and does not do.}
\(\mathcal{L}_{\mathrm{jac}}\) is a structural prior on the local gradient of the velocity field, evaluated at sampled CFM states \((z_{t,\tau},\tau)\). It does not guarantee that every multi-step autoregressive rollout under every history and plan satisfies the bounds, since that would require constraining the full forecast operator. Instead, it pushes each pointwise driver sensitivity \(J_c^{(b,t)}\), after the unit conversion in Eq.~\eqref{eq:sensitivity}, into physiologically plausible sensitivity bounds at training points. Empirically, this local constraint is sufficient to produce correctly signed and dose-bounded interventional responses, as shown in Section~\ref{sec:exp_main}.

\paragraph{Solver-depth independence.}
A computational advantage of \(\mathcal{L}_{\mathrm{jac}}\) is that it is evaluated directly on the instantaneous velocity field at sampled CFM states, rather than through the \(K_f\)-step Euler solver used during rollout. Over the full horizon, computing \(\mathcal{L}_{\mathrm{jac}}\) requires one batched velocity evaluation and one batched reverse-mode derivative with respect to the smoothed drivers, with leading cost \(\mathcal{O}(BH\,C_J)\), where \(C_J\) denotes the cost of this local Jacobian computation. By contrast, a trajectory-level sensitivity penalty would require differentiating through all \(K_f\) Euler steps, increasing compute and memory by a factor proportional to \(K_f\). Thus, the intervention regularizer is solver-free, although IFM still uses solver-based rollouts for inference and for the fidelity loss.
 
\subsection{Fidelity Loss}
\label{sec:fid}

The CFM and Jacobian terms shape the local velocity field, but they do not directly penalize errors that accumulate during the discrete autoregressive rollout. To provide global rollout supervision without the cost of full-batch solver-based training, we add a lightweight fidelity loss that compares the free-running latent forecast \(\hat z_{1:H}\) with the ground-truth latent trajectory \(z_{1:H}^{\star}\) obtained from Eq.~\eqref{eq:encode}.

For efficiency, we evaluate this loss on a random subset \(\mathcal{S}\subseteq\{1,\dots,B\}\) with \(|\mathcal{S}|=\lfloor \rho B \rfloor\), where \(\rho\in(0,1]\) and typically \(\rho\ll 1\):
\begin{equation}
    \mathcal{L}_{\mathrm{fid}}
    =
    \frac{1}{|\mathcal{S}|H}
    \sum_{i\in\mathcal{S}}\sum_{t=1}^{H}
    \left(
        \hat z_t^{(i)} - z_t^{\star\,(i)}
    \right)^2 \, .
    \label{eq:lfid}
\end{equation}
This term provides end-to-end autoregressive factual grounding while keeping the main optimization dominated by the local CFM and Jacobian objectives.



\section{Experiments}
\label{sec:experiments}

\paragraph{Dataset.}
We evaluate IFM on the simulated diabetes dataset provided
by~\citep{fathkouhi2026stationarity}. The dataset contains 100 virtual adult
subjects, each with 56 days of data generated with the UVA/Padova type 1
diabetes simulator~\citep{man2014uva}. Each trajectory is sampled every
5 minutes and contains glucose measurements, insulin delivery, and carbohydrate
intake. The simulation follows a daily meal protocol with breakfast, lunch, and
dinner occurring within prespecified time ranges, with stochastic variation in
both meal timing and meal size. No exercise covariates are present.

We use an \(80/10/10\) patient-level train/validation/test split, so test
performance is evaluated on unseen virtual subjects. Sliding-window examples
are extracted from each subject trajectory with history length \(L=2016\),
corresponding to one week of 5-minute observations, stride \(24\), and forecast
horizon \(H=24\), corresponding to 2~hours. Each example pairs the multivariate
history \(X_{1:L}\) with future drivers \(U_{1:H}\), where \(U_{1:H}\) contains
insulin and carbohydrate inputs over the prediction horizon.

\paragraph{Baselines and training.}
We compare against neural and hybrid forecasting baselines adapted from
H2NCM~\citep{zou2024hybrid}. Following H2NCM, each baseline is trained with
$
    \mathcal{L}
    =
    (1-\alpha)\mathcal{L}_{\mathrm{pred}}
    +
    \alpha\mathcal{L}_{\mathrm{rank}},
$
where \(\mathcal{L}_{\mathrm{pred}}\) fits observed-driver forecasts,
\(\mathcal{L}_{\mathrm{rank}}\) ranks the strongest-response intervention branch
highest, and \(\alpha\) controls the tradeoff. For each model family, we tune
with seed \(42\), select the configuration with the lowest validation MSE, and
rerun it with seeds \(\{1,10,42,50\}\). Main tables report seed-averaged means;
standard deviations and full sweeps are provided in Appendix~\ref{apx:full_results}.
Throughout, we fix \(\rho=0.1\), \(\lambda_{\mathrm{jac}}=1\), and
\(\lambda_{\mathrm{fid}}=1\) unless otherwise specified. Search spaces, selected
configurations, and additional baseline details are provided in
Appendix~\ref{apx:training_details} and~\ref{apx:comparison_baselines}.

\paragraph{Evaluation.}
We report observed-driver RMSE over 30, 60, and 120 minutes. Interventional
plausibility is evaluated by perturbing future insulin or carbohydrate drivers
and measuring sensitivity, strict direction, and strict ranking consistency:
\(S_{\mathrm{ins}}\), \(S_{\mathrm{carb}}\), \(D_{\mathrm{strict}}\), and
\(R_{\mathrm{strict}}\). These response-plausibility metrics, defined in
Appendix~\ref{apx:interventional_metrics}, assess magnitude, sign, and ordering
under planned driver perturbations; they are not paired interventional errors.
Qualitative 2-hour IFM interventional forecasts under planned insulin doses of
2, 4, 8, and 10 units and carbohydrate intakes of 20, 40, 80, and 100 grams are
provided in Appendix~\ref{apx:qualitative}.

\begin{table}[t]
\centering
\caption{
Representative non-degenerate points from the accuracy--intervention response frontier.
For IFM, \(\alpha=\lambda_{\mathrm{jac}}\); for baselines, \(\alpha\) is the H2NCM prediction/ranking tradeoff.
Baseline rows with \(\alpha=1\) are omitted because the prediction term is removed.
Lower RMSE is better; higher intervention metrics are better.
Full sweep results are reported in Appendix~\ref{apx:full_results}, Table~\ref{tab:alpha_sweep_selected}.
}
\label{tab:main_frontier}
\resizebox{\textwidth}{!}{
\begin{tabular}{lccccccccc}
\toprule
Model & \(\alpha\) & RMSE\(_{30}\) & RMSE\(_{60}\) & RMSE\(_{120}\)
& \(S_{\mathrm{ins}}\) & \(S_{\mathrm{carb}}\)
& \(D_{\mathrm{strict}}\) & \(R_{\mathrm{strict}}\) \\
\midrule
IFM     & 0.0 & 8.76 & 12.65 & 18.30 & 15.80& 17.08 & 0.478 & 0.002 \\
IFM     & 0.1 & 9.11 & 14.81 & 23.53 & 6.17 & 36.02 & 0.936 & 0.914 \\
IFM     & 0.3 & 9.10 & 14.34 & 23.40 & 7.49 & 39.34 & 0.937 & 0.917 \\
IFM     & 0.5 & 9.31 & 14.71 & 24.05 & 7.35 & 35.72 & 0.938 & 0.917 \\
IFM     & 0.7 & 9.01 & 14.40 & 23.91 & 7.92 & 35.23 & 0.938 & 0.918 \\
IFM     & 1.0 & 9.11 & 14.74 & 24.90 & 7.59 & 37.92 & 0.938 & 0.918 \\
\midrule
LSTM    & 0.7 & 5.01 & 8.29  & 13.11 & 3.29 & 19.35 & 0.899 & 0.803 \\
\midrule
Hovorka & 0.1 & 8.24 & 13.26 & 19.36 & 2.94 & 41.48 & 0.762 & 0.609 \\
Hovorka & 0.5 & 8.50 & 13.42 & 20.96 & 1.18 & 51.39 & 0.743 & 0.570 \\
\midrule
BNODE   & 0.1 & 8.93 & 13.17 & 18.63 & 1.67 & 23.56 & 0.883 & 0.765 \\
BNODE   & 0.7 & 9.56 & 14.38 & 20.91 & 1.14 & 35.77 & 0.941 & 0.894 \\
\midrule
S4D     & 0.1 & 13.00& 13.81 & 14.81 & 2.84 & 16.09 & 0.763 & 0.533 \\
\midrule
LP-Reduced & 0.7 & 18.73 & 24.54 & 33.28 & 0.67 & 24.75 & 0.667 & 0.497 \\
\midrule
TCN     & 0.1 & 8.13 & 9.81 & 12.06 & 9.28 & 13.90 & 0.829 & 0.660 \\
TCN     & 0.7 & 9.81 & 12.10 & 15.40 & 8.31 & 29.58 & 0.795 & 0.589 \\
\bottomrule
\end{tabular}}
\end{table}

\subsection{Forecasting Accuracy and Interventional Response}
\label{sec:exp_main}

Table~\ref{tab:main_frontier} summarizes representative non-degenerate points from the accuracy--intervention response frontier; the full sweep is reported in Appendix~\ref{apx:full_results}, Table~\ref{tab:alpha_sweep_selected}. For IFM, \(\alpha\) denotes the Jacobian weight \(\lambda_{\mathrm{jac}}\); for the H2NCM-style baselines, it denotes the prediction/ranking tradeoff in \((1-\alpha)\mathcal{L}_{\mathrm{pred}}+\alpha\mathcal{L}_{\mathrm{rank}}\). Because \(\alpha=1\) removes the prediction term for the baselines, we exclude those rows from accuracy--response tradeoff claims and treat them only as response-only endpoints.

\noindent\textbf{IFM achieves strong response magnitude and intervention validity.}
Across nonzero Jacobian weights, IFM maintains strong responses to both drivers, with \(S_{\mathrm{ins}}\in[6.17,7.92]\) and \(S_{\mathrm{carb}}\in[35.23,39.34]\), while also achieving high strict validity, \(D_{\mathrm{strict}}\in[0.936,0.938]\) and \(R_{\mathrm{strict}}\in[0.914,0.918]\). No baseline row with \(\alpha<1\) matches this combination. Hovorka responds strongly to carbohydrates, \(S_{\mathrm{carb}}\in[41.48,51.39]\), but weakly to insulin, \(S_{\mathrm{ins}}\leq2.94\). BNODE achieves high ranking consistency, up to \(D_{\mathrm{strict}}=0.941\) and \(R_{\mathrm{strict}}=0.894\), but has weak insulin response, \(S_{\mathrm{ins}}\leq1.67\). LP-Reduced has weak insulin response, \(S_{\mathrm{ins}}\leq0.67\), and lower ranking consistency, \(R_{\mathrm{strict}}\leq0.497\), across \(\alpha<1\). LSTM has the lowest RMSE, but weaker carbohydrate response, \(S_{\mathrm{carb}}\leq19.35\), and lower ranking consistency, \(R_{\mathrm{strict}}\leq0.803\). TCN is the closest in channel sensitivity, with \(S_{\mathrm{ins}}\leq9.28\) and \(S_{\mathrm{carb}}\leq29.58\), but has lower ranking consistency, \(R_{\mathrm{strict}}\leq0.66\).

\noindent\textbf{The accuracy cost remains controlled.}
Adding the Jacobian penalty sacrifices some observed-driver accuracy relative to \(\alpha=0\), but does not cause an RMSE collapse. Across nonzero \(\lambda_{\mathrm{jac}}\), RMSE remains in a narrow range: RMSE$_{30}\in[9.01,9.31]$, RMSE$_{60}\in[14.34,14.81]$, and RMSE$_{120}\in[23.40,24.90]$. Thus, IFM trades a moderate increase in RMSE for substantially improved intervention validity while keeping CFM and fidelity supervision active.

\begin{table}[t]
\centering
\caption{
Representative fidelity ablations for IFM. The \(\rho\) rows vary the fidelity subsampling ratio, while the \(\lambda_{\mathrm{fid}}\) rows vary the fidelity-loss weight.
Lower RMSE is better; higher intervention metrics are better.
Full sweeps are reported in Appendix~\ref{apx:full_results}, Tables~\ref{tab:fidelity_subsample_ratio_sweep_selected} and~\ref{tab:lambda_fidelity_sweep_selected}.
}
\label{tab:fidelity_ablation_small}
\resizebox{\textwidth}{!}{
\begin{tabular}{llrrrrrrr}
\toprule
Sweep & Setting & RMSE$_{30}$ & RMSE$_{60}$ & RMSE$_{120}$ 
& $S_{\mathrm{ins}}$ & $S_{\mathrm{carb}}$ & $D_{\mathrm{strict}}$ & $R_{\mathrm{strict}}$ \\
\midrule
\(\rho\) & \(0.1\) & 8.93 & 14.26 & 23.32 & 7.26 & 39.89 & 0.936 & 0.915 \\
\(\rho\) & \(0.7\) & 8.41 & 13.09 & 21.16 & 6.63 & 40.39 & 0.928 & 0.898 \\
\midrule
\(\lambda_{\mathrm{fid}}\) & \(0.0\) & 11.66 & 25.16 & 52.10 & 29.82 & 55.37 & 0.935 & 0.913 \\
\(\lambda_{\mathrm{fid}}\) & \(0.1\) & 9.62 & 15.48 & 26.31 & 11.59 & 42.26 & 0.934 & 0.910\\
\(\lambda_{\mathrm{fid}}\) & \(0.5\) & 9.31 & 15.34 & 25.45 & 8.34 & 36.53 & 0.935 & 0.912 \\
\(\lambda_{\mathrm{fid}}\) & \(1.0\) & 9.24 & 14.24 & 23.71 & 7.25 & 36.22 & 0.938 & 0.918 \\
\bottomrule
\end{tabular}}
\end{table}

\subsection{Ablations}
\label{sec:exp_ablation}

\paragraph{Effect of removing the Jacobian penalty.}
The \(\lambda_{\mathrm{jac}}=\alpha=0\) IFM row in Table~\ref{tab:main_frontier} isolates the model without Jacobian regularization. Although this setting gives the best IFM observed-driver RMSE, \(8.76/12.65/18.30\) mg/dL at \(30/60/120\) minutes, its intervention behavior is not physiologically reliable: \(D_{\mathrm{strict}}=0.478\) and \(R_{\mathrm{strict}}=0.002\), with weak carbohydrate response \(S_{\mathrm{carb}}=17.08\), despite large insulin sensitivity \(S_{\mathrm{ins}}=15.80\). This shows that strong observational accuracy does not imply valid prospective intervention response. 

\paragraph{Fidelity ablation.}
Table~\ref{tab:fidelity_ablation_small} shows that the fidelity term mainly improves autoregressive accuracy without removing the intervention behavior induced by the Jacobian regularizer. Increasing the subsampling ratio \(\rho\) improves long-horizon RMSE, with RMSE$_{120}$ decreasing from \(23.32\) at \(\rho=0.1\) to \(21.16\) at \(\rho=0.7\), while directional consistency remains high, \(D_{\mathrm{strict}}\in[0.928,0.936]\). Even the default lightweight setting \(\rho=0.1\) remains effective, preserving strong intervention response while maintaining competitive observed-driver accuracy. The \(\lambda_{\mathrm{fid}}\) sweep further shows that fidelity supervision is necessary: without it, RMSE degrades to \(11.66/25.16/52.10\) at \(30/60/120\) minutes, whereas \(\lambda_{\mathrm{fid}}=1.0\) improves RMSE to \(9.24/14.24/23.71\). For all \(\lambda_{\mathrm{fid}}>0\), intervention validity remains stable, with \(D_{\mathrm{strict}}\in[0.934,0.938]\) and \(R_{\mathrm{strict}}\in[0.91,0.918]\).

Additional ablations on physiological sensitivity bounds are provided in Appendix~\ref{apx:local_bounds}.

\section{Discussion}
\label{sec:discussion}

IFM separates forecasting from intervention-response control: the encoder and flow model learn glucose dynamics from history, while the Jacobian penalty constrains local responses to planned insulin and carbohydrate inputs. This places IFM between unconstrained sequence models, which may learn confounded driver effects, and mechanistic ODE models, which require fixed physiological scaffolds.

The Jacobian penalty is a structural prior, not a causal identification guarantee. It constrains local effective-driver sensitivities but does not recover arbitrary interventional distributions under unmeasured confounding. Our goal is prospective response regularization: given credible directional priors, IFM encourages physiologically valid local responses and tests whether they persist through autoregressive rollout. Driver isolation helps prevent interventions from bypassing the regularized velocity field, although extreme sparsity or confounding may still cause treatment underuse, as shown in Appendix~\ref{apx:pk_smooth}. This principle may extend to other dose--response settings with sparse interventions, delayed effects, and calibrated sensitivity bounds.

No single interventional metric is sufficient. Sensitivity captures response
magnitude, while \(D_{\mathrm{strict}}\) and \(R_{\mathrm{strict}}\) capture
sign and ordering. Thus, a model can rank interventions correctly but respond
too weakly, or show large responses with invalid direction. We therefore compare
models using all three aspects jointly.
\section{Limitations}
\label{sec:limitations}

IFM has several limitations. Its Jacobian bounds are fixed at the population level and do not capture patient-specific sensitivities. Experiments use a single simulated UVA/Padova cohort~\citep{fathkouhi2026stationarity}; validation on real clinical data remains open, especially with missing or noisy carbohydrate records. The method is only partially solver-free, since the fidelity loss still uses subsampled rollouts. Finally, IFM forecasts glucose under candidate plans but does not prescribe doses. Future work should study patient-adaptive bounds, uncertainty calibration, real-world validation, and integration with safety-constrained decision support.

\section{Conclusion}
\label{sec:conclusion}
We introduced Interventional Flow Matching (IFM), a conditional generative framework for prospective clinical dose--response forecasting. IFM constrains intervention semantics at the instantaneous velocity-field level: its Jacobian regularizer enforces signed, dose-bounded local responses to planned insulin and carbohydrate drivers using directional physiological priors. IFM still uses rollout simulation for forecasting and fidelity supervision; the physiological response constraint itself is imposed locally on the velocity field.

In physiological glucose forecasting, IFM maintains competitive observed-driver
RMSE and improves interventional validity, producing physiologically meaningful
insulin-lowering and carbohydrate-raising responses with high directional and
ranking consistency. More broadly, these results suggest that local response-level regularization can complement learned rollout forecasting when reliable directional priors are available.


\begin{ack}
This work was supported by the Launchpad Foundation.
\end{ack}

\bibliographystyle{plain}
\bibliography{references}

@article{gu2023mamba,
  title={Mamba: Linear-time sequence modeling with selective state spaces},
  author={Gu, Albert and Dao, Tri},
  journal={arXiv preprint arXiv:2312.00752},
  year={2023}
}

@article{zhang2019root,
  title={Root mean square layer normalization},
  author={Zhang, Biao and Sennrich, Rico},
  journal={Advances in Neural Information Processing Systems},
  volume={32},
  year={2019}
}

@inproceedings{zhang2025dim,
  title={DiM-Gestor: Co-Speech Gesture Generation with Adaptive Layer Normalization Mamba-2},
  author={Zhang, Fan and Wei, Yi and Ji, Naye and Wu, Jingmei and Wang, Zhaohan and Gao, Fuxing and Zhang, Liuqing and Ye, Zhenqing and Yan, Leyao and Dai, Lanxin and others},
  booktitle={2025 International Conference on Computer Vision, Image Processing and Computational Photography (CVIP)},
  pages={01--13},
  year={2025},
  organization={IEEE}
}

@inproceedings{lipman2023flow,
  title={Flow Matching for Generative Modeling},
  author={Lipman, Yaron and Chen, Ricky TQ and Ben-Hamu, Heli and Nickel, Maximilian and Le, Matt},
  booktitle={11th International Conference on Learning Representations, ICLR 2023},
  year={2023}
}

@book{pearl2009causality,
  title={Causality},
  author={Pearl, Judea},
  year={2009},
  publisher={Cambridge University Press}
}

@inproceedings{wu2025doflow,
  title={Doflow: Flow-based generative models for interventional and counterfactual forecasting on time series},
  author={Wu, Dongze and Qiu, Feng and Xie, Yao},
  booktitle={The 14th International Conference on Learning Representations},
  year={2025}
}

@inproceedings{zou2024hybrid,
  title={Hybrid $^2$ Neural ODE Causal Modeling and an Application to Glycemic Response},
  author={Zou, Bob Junyi and Levine, Matthew E and Zaharieva, Dessi P and Johari, Ramesh and Fox, Emily},
  booktitle={International Conference on Machine Learning},
  pages={62934--62963},
  year={2024},
  organization={PMLR}
}

@article{hovorka2004nonlinear,
  title={Nonlinear model predictive control of glucose concentration in subjects with type 1 diabetes},
  author={Hovorka, Roman and Canonico, Valentina and Chassin, Ludovic J and Haueter, Ulrich and Massi-Benedetti, Massimo and Federici, Marco Orsini and Pieber, Thomas R and Schaller, Helga C and Schaupp, Lukas and Vering, Thomas and others},
  journal={Physiological Measurement},
  volume={25},
  number={4},
  pages={905--920},
  year={2004}
}

@article{gu2022parameterization,
  title={On the parameterization and initialization of diagonal state space models},
  author={Gu, Albert and Goel, Karan and Gupta, Ankit and R{\'e}, Christopher},
  journal={Advances in Neural Information Processing Systems},
  volume={35},
  pages={35971--35983},
  year={2022}
}

@inproceedings{lea2016temporal,
  title={Temporal convolutional networks: A unified approach to action segmentation},
  author={Lea, Colin and Vidal, Rene and Reiter, Austin and Hager, Gregory D},
  booktitle={European Conference on Computer Vision},
  pages={47--54},
  year={2016},
  organization={Springer}
}

@article{fathkouhi2026stationarity,
  title={The Stationarity Bias: Stratified Stress-Testing for Time-Series Imputation in Regulated Dynamical Systems},
  author={Fathkouhi, Amirreza Dolatpour and Namazi, Alireza and Shakeri, Heman},
  journal={arXiv preprint arXiv:2602.15637},
  year={2026}
}

@inproceedings{Loshchilov2017DecoupledWD,
  title={Decoupled Weight Decay Regularization},
  author={Ilya Loshchilov and Frank Hutter},
  booktitle={International Conference on Learning Representations},
  year={2017}
}

@inproceedings{zou2026automatic,
  title={Automatic and Structure-Aware Sparsification of Hybrid Neural {ODE}s with Application to Glucose Prediction},
  author={Zou, Bob Junyi and Tian, Lu},
  booktitle={The 14th International Conference on Learning Representations},
  year={2026}
}

@inproceedings{bica2020estimating,
  title     = {Estimating Counterfactual Treatment Outcomes over Time through Adversarially Balanced Representations},
  author    = {Bica, Ioana and Alaa, Ahmed M. and Jordon, James and van der Schaar, Mihaela},
  booktitle = {International Conference on Learning Representations},
  year      = {2020},
  url       = {https://openreview.net/forum?id=BJg866NFvB}
}

@inproceedings{li2021gnet,
  title     = {{G}-Net: A Recurrent Network Approach to {G}-Computation for Counterfactual Prediction under a Dynamic Treatment Regime},
  author    = {Li, Rui and Hu, Stephanie and Lu, Mingyu and Utsumi, Yuria and Chakraborty, Prithwish and Sow, Daby M. and Madan, Piyush and Li, Jun and Ghalwash, Mohamed and Shahn, Zach and Lehman, Li-wei},
  booktitle = {Proceedings of Machine Learning for Health},
  pages     = {282--299},
  year      = {2021},
  editor    = {Roy, Subhrajit and Pfohl, Stephen and Rocheteau, Emma and Tadesse, Girmaw Abebe and Oala, Luis and Falck, Fabian and Zhou, Yuyin and Shen, Liyue and Zamzmi, Ghada and Mugambi, Purity and Zirikly, Ayah and McDermott, Matthew B. A. and Alsentzer, Emily},
  volume    = {158},
  series    = {Proceedings of Machine Learning Research},
  publisher = {PMLR},
  url       = {https://proceedings.mlr.press/v158/li21a.html}
}

@inproceedings{melnychuk2022causal,
  title     = {Causal Transformer for Estimating Counterfactual Outcomes},
  author    = {Melnychuk, Valentyn and Frauen, Dennis and Feuerriegel, Stefan},
  booktitle = {Proceedings of the 39th International Conference on Machine Learning},
  pages     = {15293--15329},
  year      = {2022},
  editor    = {Chaudhuri, Kamalika and Jegelka, Stefanie and Song, Le and Szepesvari, Csaba and Niu, Gang and Sabato, Sivan},
  volume    = {162},
  series    = {Proceedings of Machine Learning Research},
  publisher = {PMLR},
  url       = {https://proceedings.mlr.press/v162/melnychuk22a.html}
}

@article{man2014uva,
  title={The {UVA}/{PADOVA} type 1 diabetes simulator: new features},
  author={Man, Chiara Dalla and Micheletto, Francesco and Lv, Dayu and Breton, Marc and Kovatchev, Boris and Cobelli, Claudio},
  journal={Journal of Diabetes Science and Technology},
  volume={8},
  number={1},
  pages={26--34},
  year={2014},
  publisher={SAGE Publications Sage CA: Los Angeles, CA}
}

@misc{dalla2009physical,
  title={Physical activity into the meal glucose—Insulin model of type 1 diabetes: In silico studies},
  author={Dalla Man, Chiara and Breton, Marc D and Cobelli, Claudio},
  year={2009},
  publisher={SAGE Publications Sage CA: Los Angeles, CA}
}

@article{shakeri2025driver,
  title={The Driver-Blindness Phenomenon: Why Deep Sequence Models Default to Autocorrelation in Blood Glucose Forecasting},
  author={Shakeri, Heman},
  journal={arXiv preprint arXiv:2511.20601},
  year={2025}
}

@article{tancik2020fourier,
  title={Fourier features let networks learn high frequency functions in low dimensional domains},
  author={Tancik, Matthew and Srinivasan, Pratul and Mildenhall, Ben and Fridovich-Keil, Sara and Raghavan, Nithin and Singhal, Utkarsh and Ramamoorthi, Ravi and Barron, Jonathan and Ng, Ren},
  journal={Advances in Neural Information Processing Systems},
  volume={33},
  pages={7537--7547},
  year={2020}
}

@inproceedings{xia2023neural,
  title={Neural Causal Models for Counterfactual Identification and Estimation},
  author={Xia, Kevin Muyuan and Pan, Yushu and Bareinboim, Elias},
  booktitle={11th International Conference on Learning Representations, ICLR 2023},
  year={2023}
}

@inproceedings{wu2022learning,
  title={Learning large-scale subsurface simulations with a hybrid graph network simulator},
  author={Wu, Tailin and Wang, Qinchen and Zhang, Yinan and Ying, Rex and Cao, Kaidi and Sosic, Rok and Jalali, Ridwan and Hamam, Hassan and Maucec, Marko and Leskovec, Jure},
  booktitle={Proceedings of the 28th ACM SIGKDD Conference on Knowledge Discovery and Data Mining},
  pages={4184--4194},
  year={2022}
}

@inproceedings{allen2023graph,
  title={Graph network simulators can learn discontinuous, rigid contact dynamics},
  author={Allen, Kelsey R and Guevara, Tatiana Lopez and Rubanova, Yulia and Stachenfeld, Kim and Sanchez-Gonzalez, Alvaro and Battaglia, Peter and Pfaff, Tobias},
  booktitle={Conference on Robot Learning},
  pages={1157--1167},
  year={2023},
  organization={PMLR}
}

@article{li2022graph,
  title={Graph neural networks accelerated molecular dynamics},
  author={Li, Zijie and Meidani, Kazem and Yadav, Prakarsh and Barati Farimani, Amir},
  journal={The Journal of Chemical Physics},
  volume={156},
  number={14},
  year={2022},
  publisher={AIP Publishing}
}

@inproceedings{miller2020learning,
  title={Learning Insulin-Glucose Dynamics in the Wild},
  author={Miller, Andrew C and Foti, Nicholas J and Fox, Emily},
  booktitle={Machine Learning for Healthcare Conference},
  pages={172--197},
  year={2020},
  organization={PMLR}
}

@inproceedings{tamir2024conditional,
  title={Conditional flow matching for time series modelling},
  author={Tamir, Ella and Laabid, Najwa and Heinonen, Markus and Garg, Vikas and Solin, Arno},
  booktitle={ICML 2024 Workshop on Structured Probabilistic Inference \& Generative Modeling},
  year={2024}
}

@inproceedings{scassola2026graph,
  title={Graph-conditional flow matching for relational data generation},
  author={Scassola, Davide and Saccani, Sebastiano and Bortolussi, Luca},
  booktitle={Proceedings of the AAAI Conference on Artificial Intelligence},
  volume={40},
  number={30},
  pages={25209--25217},
  year={2026}
}

@article{bahadori2017causal,
  title={Causal regularization},
  author={Bahadori, Mohammad Taha and Chalupka, Krzysztof and Choi, Edward and Chen, Robert and Stewart, Walter F and Sun, Jimeng},
  journal={arXiv preprint arXiv:1702.02604},
  year={2017}
}

@inproceedings{nie2021vcnet,
  title={{VCN}et and Functional Targeted Regularization For Learning Causal Effects of Continuous Treatments},
  author={Nie, Lizhen and Ye, Mao and Liu, Qiang and Nicolae, Dan},
  booktitle={International Conference on Learning Representations},
  year={2021}
}

@article{dong2024background,
  title={Background debiased {SAR} automatic target recognition via a novel causal interventional regularizer},
  author={Dong, Hongwei and Han, Fangzhou and Si, Lingyu and Qiang, Wenwen and Zhang, Ruiheng and Zhang, Lamei},
  journal={IEEE Journal of Selected Topics in Applied Earth Observations and Remote Sensing},
  volume={17},
  pages={16993--17006},
  year={2024},
  publisher={IEEE}
}

@inproceedings{pfaff2021learning,
  title={Learning Mesh-Based Simulation with Graph Networks},
  author={Pfaff, Tobias and Fortunato, Meire and Sanchez-Gonzalez, Alvaro and Battaglia, Peter W},
  booktitle={International Conference on Learning Representations},
  year={2021}
}

@article{lee2024shortcomings,
  title={Shortcomings in the evaluation of blood glucose forecasting},
  author={Lee, Jung Min and Pop-Busui, Rodica and Lee, Joyce M and Fleischer, Jesper and Wiens, Jenna},
  journal={IEEE Transactions on Biomedical Engineering},
  volume={71},
  number={12},
  pages={3424--3431},
  year={2024},
  publisher={IEEE}
}

\appendix
\section{Architecture Details}
\label{apx:arch}
\subsection{Encoder Architecture}
\label{apx:encoder}

The encoder receives
\(X_{1:L}\in\mathbb{R}^{B\times L\times d_{\mathrm{in}}}\) and projects each
time step to the model dimension:
\begin{align}
    R_{1:L}^{(0)}
        &= \mathrm{Linear}_{d_{\mathrm{in}}\rightarrow d_m}(X_{1:L}) \, ,
        &
        R_{1:L}^{(0)}\in\mathbb{R}^{B\times L\times d_m} \, .
\end{align}
The projected sequence is processed by \(L_e\) residual Mamba~\citep{gu2023mamba} blocks with
RMSNorm\citep{zhang2019root} pre-normalization:
\begin{align}
    \tilde R_{1:L}^{(\ell-1)}
        &= \mathrm{RMSNorm}\!\left(R_{1:L}^{(\ell-1)}\right) \, , \\
    R_{1:L}^{(\ell)}
        &= R_{1:L}^{(\ell-1)}
        + \mathrm{Mamba}_{\ell}\!\left(\tilde R_{1:L}^{(\ell-1)}\right) \, ,
        &
        \ell=1,\dots,L_e \, , \\
    \tilde{R}_{1:L}^{L_e}
        &= \mathrm{RMSNorm}\!\left(R_{1:L}^{(L_e)}\right) \, ,
        &
        \tilde{R}_{1:L}^{L_e}\in\mathbb{R}^{B\times L\times d_m} \, .
\end{align}
The initial recurrent state is the final encoded history state:
\begin{align}
    h_0 = \tilde{R}_{L}^{L_e} \in\mathbb{R}^{B\times d_m} \, .
\end{align}

\subsection{Physiological Smoothing of Future Drivers}
\label{apx:pksmooth}

Insulin and carbohydrate interventions arrive as sparse impulses, whereas their
physiological effects are delayed and distributed over time. IFM therefore maps
the planned driver sequence
\(U_{1:H}\in\mathbb{R}^{B\times H\times d_u}\) to an effective smoothed sequence
\(D_{1:H}=(d_1,\dots,d_H)\in\mathbb{R}^{B\times H\times d_u}\).

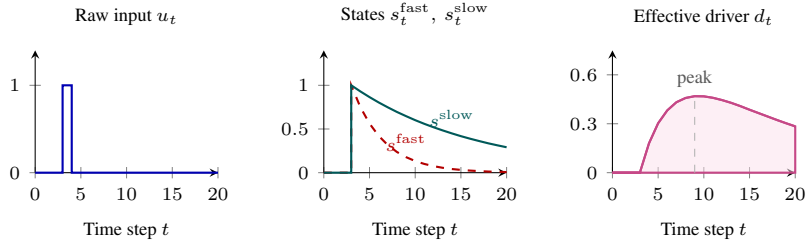
\begin{figure}[H]
  \centering
  \begin{tikzpicture}
\begin{groupplot}[
    group style={
        group size=3 by 1,
        horizontal sep=1.4cm,
    },
    width=4.0cm, height=3.2cm,
    xlabel style={font=\scriptsize},
    ylabel style={font=\scriptsize},
    tick label style={font=\scriptsize},
    xmin=0, xmax=20,
    ymin=0,
    axis lines=left,
    xtick={0,5,10,15,20},
    xlabel={Time step $t$},
    every axis plot/.append style={thick},
    clip=false,
]

\nextgroupplot[
    title={\scriptsize Raw input $u_t$},
    ymax=1.4,
    ytick={0,1},
]
\addplot[blue!70!black, const plot] coordinates {
    (0,0)(3,1)(4,0)(20,0)
};

\nextgroupplot[
    title={\scriptsize States $s_t^{\mathrm{fast}},\, s_t^{\mathrm{slow}}$},
    ymax=1.4,
    ytick={0,0.5,1},
]
\addplot[red!70!black, dashed] coordinates {
    (0,0)(3,0)
    (3,1)(4,0.75)(5,0.5625)(6,0.422)(7,0.316)(8,0.237)
    (9,0.178)(10,0.133)(11,0.100)(12,0.075)(13,0.056)
    (14,0.042)(15,0.032)(16,0.024)(17,0.018)(18,0.013)
    (19,0.010)(20,0.007)
};
\addplot[teal!70!black] coordinates {
    (0,0)(3,0)
    (3,1)(4,0.93)(5,0.865)(6,0.804)(7,0.748)(8,0.695)
    (9,0.647)(10,0.601)(11,0.559)(12,0.520)(13,0.484)
    (14,0.450)(15,0.418)(16,0.389)(17,0.362)(18,0.336)
    (19,0.313)(20,0.291)
};
\node[font=\scriptsize, red!70!black]  at (axis cs:9,0.35)  {$s^{\mathrm{fast}}$};
\node[font=\scriptsize, teal!70!black] at (axis cs:14,0.62) {$s^{\mathrm{slow}}$};

\nextgroupplot[
    title={\scriptsize Effective driver $d_t$},
    ymax=0.75,
    ytick={0,0.3,0.6},
]
\addplot[magenta!80!black, fill=magenta!15, fill opacity=0.5] coordinates {
    (0,0)(3,0)
    (3,0)(4,0.18)(5,0.303)(6,0.382)(7,0.432)(8,0.458)
    (9,0.469)(10,0.468)(11,0.459)(12,0.445)(13,0.428)
    (14,0.408)(15,0.386)(16,0.365)(17,0.344)(18,0.323)
    (19,0.303)(20,0.284)
} \closedcycle;
\addplot[magenta!80!black] coordinates {
    (0,0)(3,0)
    (3,0)(4,0.18)(5,0.303)(6,0.382)(7,0.432)(8,0.458)
    (9,0.469)(10,0.468)(11,0.459)(12,0.445)(13,0.428)
    (14,0.408)(15,0.386)(16,0.365)(17,0.344)(18,0.323)
    (19,0.303)(20,0.284)
};
\draw[dashed, gray!60] (axis cs:9,0) -- (axis cs:9,0.469)
    node[above, font=\scriptsize, gray!70!black] {peak};

\end{groupplot}
\end{tikzpicture}
  \caption{
  Physiological smoothing of a single unit impulse at \(t=3\).
  Left: raw input \(u_t\). Middle: fast and slow decay states.
  Right: effective driver
  \(d_t=\eta\odot\operatorname{ReLU}(s_t^{\mathrm{slow}}-s_t^{\mathrm{fast}})\),
  which peaks after the impulse and decays smoothly, mimicking delayed insulin or carbohydrate absorption.
  }
  \label{fig:pk_smooth}
\end{figure}

At each step \(t\), the raw driver \(u_t\in\mathbb{R}^{B\times d_u}\) is passed
through fast and slow decay states
\(s_t^{\mathrm{fast}},s_t^{\mathrm{slow}}\in\mathbb{R}^{B\times d_u}\),
initialized as \(s_0^{\mathrm{fast}}=s_0^{\mathrm{slow}}=0\):
\begin{align}
    s_t^{\mathrm{fast}} &= a_f \odot s_{t-1}^{\mathrm{fast}} + u_t \, , \\
    s_t^{\mathrm{slow}} &= a_s \odot s_{t-1}^{\mathrm{slow}} + u_t \, ,
\end{align}
where \(a_f,a_s\in(0,1)^{d_u}\) are learned per-channel decay vectors
parameterized to satisfy \(a_f<a_s\). The effective driver is
\begin{align}
    d_t = \eta \odot \operatorname{ReLU}\!\left(s_t^{\mathrm{slow}} - s_t^{\mathrm{fast}}\right) \, ,
\end{align}
where \(\eta\in\mathbb{R}_{>0}^{d_u}\) is a learned per-channel gain.
The difference between the slow and fast states produces a delayed-peak
response, illustrated in Figure~\ref{fig:pk_smooth}.

\subsection{Interventional Rollout Algorithm}
\label{apx:inference}

Algorithm~\ref{alg:interventional} summarizes the rollout procedure used for prospective intervention queries. For paired comparisons, we reuse the same base-noise samples across candidate plans so that trajectory differences reflect the planned drivers rather than independent sampling noise.

\begin{algorithm}[t]
\caption{Prospective Interventional Glucose Forecasting}
\label{alg:interventional}
\begin{algorithmic}[1]
\Require History \(X_{1:L}\), planned drivers \(U_{1:H}^{\mathrm{int}}\), flow steps \(K_f\), skew \(\kappa\), bounds \([g_{\min}, g_{\max}]\)

\State \(\tilde{R}_{1:L}^{L_e}\leftarrow \mathrm{MambaEncoder}(X_{1:L})\) \Comment{Section~\ref{sec:encoder}}
\State \(h_0\leftarrow \tilde{R}_{L}^{L_e}\)
\State \(D_{1:H}^{\mathrm{int}}=(d_1^{\mathrm{int}},\dots,d_H^{\mathrm{int}})
       \leftarrow \mathrm{PKSmooth}(U_{1:H}^{\mathrm{int}})\) \Comment{Section~\ref{sec:pksmooth}}

\For{\(t=1,\dots,H\)}
    \State \(h_{t-1}^{\mathrm{cond}}\leftarrow [h_{t-1};d_t^{\mathrm{int}}]\)
    \State \(z_{t,0}^{\mathrm{int}}\sim\mathcal{N}(0,I)\)
    \For{\(k=0,\dots,K_f-1\)} \Comment{Euler steps (Eq.~\ref{eq:euler})}
        \State \(\tau_k\leftarrow k/K_f\)
        \State \(z_{t,\tau_{k+1}}^{\mathrm{int}}
        \leftarrow
        z_{t,\tau_k}^{\mathrm{int}}
        +
        \frac{1}{K_f}
        v_\theta\!\left(
            h_{t-1}^{\mathrm{cond}},
            z_{t,\tau_k}^{\mathrm{int}},
            \tau_k
        \right)\)
    \EndFor
    \State \(\hat z_t^{\mathrm{int}}\leftarrow z_{t,\tau_{K_f}}^{\mathrm{int}}\)
    \State \(\hat G_t^{\mathrm{int}}\leftarrow
        g_{\min}
        +(g_{\max}-g_{\min})
        \sigma(\hat z_t^{\mathrm{int}})^{1/\kappa}\) \Comment{Decode latent state (Eq.~\ref{eq:decode})}
    \State \(h_t\leftarrow \mathrm{GRUCell}(\hat z_t^{\mathrm{int}},h_{t-1})\) \Comment{Progress physical time (Eq.~\ref{eq:gru})}
\EndFor

\State \Return \(\hat G_{1:H}^{\mathrm{int}}\)
\end{algorithmic}
\end{algorithm}

\section{Interventional Forecasting Evaluation Metrics}
\label{apx:interventional_metrics}

\paragraph{Notation.}
Let $i\in\{1,\dots,N\}$ index samples and $t\in\{1,\dots,H\}$ index forecast
steps. We evaluate three forecast branches per sample. Let $F_{i,t}$ denote
the observed-driver forecast, i.e., the prediction using the future drivers
observed in the data. Let $I_{i,t}$ denote the prediction obtained after
adding 2 units of insulin to the observed drivers, and let $C_{i,t}$ denote
the prediction obtained after adding 50 grams of carbohydrate to the observed
drivers. Both interventions occur at time \(t = 1\). We define interventional effects relative to the observed-driver
forecast:
\begin{equation}
    \Delta^I_{i,t}=I_{i,t}-F_{i,t} \, ,
    \qquad
    \Delta^C_{i,t}=C_{i,t}-F_{i,t} \, .
\end{equation}
Physiologically consistent responses have
$\Delta_{i,t}^I < 0$
and
$\Delta^C_{i,t}>0$.

\paragraph{Observed-driver forecasting error (RMSE).}
For the observed-driver forecast, prediction error is measured against the
observed glucose trajectory $G_{i,t}$:
\begin{equation}
    \mathrm{RMSE}
    =
    \sqrt{
    \frac{1}{NH}
    \sum_{i=1}^{N}\sum_{t=1}^{H}
    \left(F_{i,t}-G_{i,t}\right)^2
    } \, .
    \label{eq:eval_rmse}
\end{equation}

\paragraph{Boundary-aware masks.}
To avoid evaluating intervention effects at saturated physiological
boundaries, we mask insulin effects near the lower bound and carbohydrate
effects near the upper bound:
\begin{equation}
    M^I_{i,t}=\mathbf{1}(F_{i,t}>g_{\min}+\delta) \, ,
    \qquad
    M^C_{i,t}=\mathbf{1}(F_{i,t}<g_{\max}-\delta) \, ,
\end{equation}
with ranking mask
\begin{equation}
    M^R_{i,t}=M^I_{i,t}M^C_{i,t} \, .
\end{equation}
We use \(\delta = 10\) mg/dL.

\paragraph{Masked interventional sensitivity ($S_{\mathrm{ins}}$, $S_{\mathrm{carb}}$).}
Sensitivity is the average absolute response to each added intervention over
its valid mask:
\begin{equation}
    S_{\mathrm{ins}}
    =
    \frac{
        \sum_{i,t}M^I_{i,t}| \Delta_{i, t}^I |
    }{
        \sum_{i,t}M^I_{i,t}
    } \, ,
    \qquad
    S_{\mathrm{carb}}
    =
    \frac{
        \sum_{i,t}M^C_{i,t}| \Delta_{i, t}^C |
    }{
        \sum_{i,t}M^C_{i,t}
    } \, .
    \label{eq:interventional_sensitivity}
\end{equation}
Sensitivity measures response magnitude, while direction is evaluated separately by \(D_{\mathrm{strict}}\).

\paragraph{Evaluable set ($\mathcal{E}$).}
Strict consistency scores are computed only on samples with valid timesteps
and non-negligible responses to both added interventions. With effect margin
$m=0.5$ mg/dL, define
\begin{equation}
    A^I_i=\max_{t:M^I_{i,t}=1}| \Delta_{i,t}^I | \, ,
    \qquad
    A^C_i=\max_{t:M^C_{i,t}=1}| \Delta_{i,t}^C | \, .
\end{equation}
The evaluable set is
\begin{equation}
    \mathcal{E}
    =
    \left\{
    i:
    \sum_t M^I_{i,t}>0,\;
    \sum_t M^C_{i,t}>0,\;
    \sum_t M^R_{i,t}>0,\;
    A^I_i>m,\;
    A^C_i>m
    \right\} \, .
    \label{eq:evaluable_interventional}
\end{equation}
All strict consistency metrics below are averaged over $\mathcal{E}$.

\paragraph{Strict directional consistency ($D_{\mathrm{strict}}$).}
For each $i\in\mathcal{E}$, directional consistency is the fraction of valid
steps for which each added intervention has the expected signed effect:
\begin{equation}
    D^I_i
    =
    \frac{
        \sum_t M^I_{i,t}\mathbf{1}( \Delta_{i,t}^I < -m)
    }{
        \sum_t M^I_{i,t}
    } \, ,
    \qquad
    D^C_i
    =
    \frac{
        \sum_t M^C_{i,t}\mathbf{1}( \Delta_{i,t}^C > m )
    }{
        \sum_t M^C_{i,t}
    } \, .
\end{equation}
The strict directional consistency score is
\begin{equation}
    D_{\mathrm{strict}}
    =
    \frac{1}{|\mathcal{E}|}
    \sum_{i\in\mathcal{E}}
    \frac{D^I_i+D^C_i}{2} \, .
    \label{eq:interventional_directional_consistency}
\end{equation}

\paragraph{Strict ranking consistency ($R_{\mathrm{strict}}$).}
Across the three branches, the expected pointwise ordering is
\begin{equation}
    I_{i,t} < F_{i,t} < C_{i,t} \, .
\end{equation}
Using margin $m$, the per-sample ranking score is
\begin{equation}
    R_i
    =
    \frac{
        \sum_t M^R_{i,t}
        \mathbf{1}( \Delta_{i,t}^I < -m )
        \mathbf{1}( \Delta_{i,t}^C > m )
    }{
        \sum_t M^R_{i,t}
    } \, .
\end{equation}
The strict ranking consistency is
\begin{equation}
    R_{\mathrm{strict}}
    =
    \frac{1}{|\mathcal{E}|}
    \sum_{i\in\mathcal{E}}R_i \, .
    \label{eq:interventional_ranking_consistency}
\end{equation}

\section{Comparison Baselines}
\label{apx:comparison_baselines}
\subsection{H2NCM Baselines}
We compare against the baseline families considered in H2NCM~\citep{zou2024hybrid}:
BNODE, LP-Reduced, LSTM, S4D~\citep{gu2022parameterization}, and TCN~\citep{lea2016temporal}. BNODE is the black-box neural ODE
baseline, which parameterizes the state dynamics entirely with a neural network.
LP-Reduced is the reduced latent-parameter hybrid model from H2NCM, retaining
the reduced UVA/Padova mechanistic dynamics~\citep{man2014uva} while allowing selected simulator
parameters to vary over time through learned latent dynamics.

We do not directly instantiate the MNODE variant from H2NCM. In that work,
MNODE is constructed from the reduced UVA/Padova mechanistic dependency graph
and includes exercise-response components~\citep{dalla2009physical} with exercise-related covariates such
as heart rate and step count. Our setting does not include exercise channels;
the available inputs are glucose, insulin, and carbohydrates. We therefore
retain the hybrid neural ODE design principle, but instantiate the mechanistic
component with the Hovorka glucose--insulin model~\citep{hovorka2004nonlinear}.
As detailed in Appendix~\ref{apx:hovorka}, the Hovorka model is driven by
insulin and carbohydrate inputs and does not require exercise modules, making
it compatible with our forecasting setting while preserving a physiologically
grounded glucose--insulin structure.

Glucose, insulin, and carbohydrate channels are
normalized using training-set statistics, and prediction accuracy is optimized
with a normalized glucose MSE loss~($\mathcal{L}_{\mathrm{pred}}$).

For the baselines, we use the driver-based intervention-ranking loss
of~\citep{zou2024hybrid}. For each training example, one intervention category is sampled uniformly,
and the ranking loss is computed from three forecasts generated by perturbing
the observed future drivers. The intervention categories follow the original
protocol: (i) carbohydrate-only, adding \(\{0,50,100\}\) grams of carbohydrate;
(ii) insulin-only, adding \(\{0,2.5,5.0\}\) units of insulin; and
(iii) mixed, using \{\text{no change}, \(+50\) grams carbohydrate,
\(+10\) units insulin\}. Each branch is scored by its mean predicted glucose
over the forecast horizon.

For branch \(k\), define
$
    \mu_k = \frac{1}{H}\sum_{t=1}^{H}\hat G_t^{(k)}.
$
Let \(r\) denote the branch that should be ranked highest under the selected
intervention category. With temperature \(\beta=10\), the ranking loss is
\begin{equation}
    \mathcal{L}_{\mathrm{rank}}
    =
    -\log
    \frac{\exp(\beta \mu_r)}
    {\sum_{j=1}^{3}\exp(\beta \mu_j)} \, .
\end{equation}
The full baseline objective is
\begin{equation}
    \mathcal{L}
    =
    (1-\alpha)\mathcal{L}_{\mathrm{pred}}
    +
    \alpha\mathcal{L}_{\mathrm{rank}} \, ,
    \qquad
    \alpha\in[0,1] \, ,
\end{equation}
where \(\alpha\) controls the trade-off between observed-driver prediction
accuracy and intervention-ranking supervision.

\subsection{Hovorka-Structured Neural Dynamics}
\label{apx:hovorka}

Following the mechanistic neural ODE (MNODE) framework~\citep{zou2024hybrid},
we model glucose evolution with a neural state-space model that preserves a
mechanistic dependency graph while learning the state transition functions from
data. We instantiate this graph using the compartmental structure of the
Hovorka glucose--insulin model~\citep{hovorka2004nonlinear}. The latent state is
\begin{equation}
\xi_t =
[Q_1,Q_2,S_1,S_2,I,x_1,x_2,x_3,G_1,G_2]_t  \, ,
\end{equation}
where $Q_1,Q_2$ are accessible and non-accessible glucose masses,
$S_1,S_2$ are subcutaneous insulin absorption compartments, $I$ is plasma
insulin, and $x_1,x_2,x_3$ are remote insulin actions on glucose transport,
disposal, and endogenous glucose production. The states $G_1,G_2$ are learned
gut absorption compartments representing the two-stage carbohydrate absorption
pathway.

Let \(\xi_{t,j}\) denote the \(j\)-th component of the latent state vector
\(\xi_t\). Each component is updated by a neural transition function that
receives only its physiologically admissible parent states and inputs:
\begin{equation}
\xi_{t+1,j}
=
\xi_{t,j}
+
\Delta t\,
f_{\theta_j}
\left(
\xi_{t,\mathrm{Pa}_\xi(j)},
u_{t,\mathrm{Pa}_u(j)}
\right) \, ,
\qquad
\Delta t=1  \, .
\end{equation}
Here \(u_t=[u^{\mathrm{ins}}_t,u^{\mathrm{carb}}_t]\) denotes insulin and
carbohydrate inputs, \(\mathrm{Pa}_\xi(j)\) and \(\mathrm{Pa}_u(j)\) are the
parent state and input sets for component \(j\), and \(f_{\theta_j}\) is the
component-specific neural transition function with learnable parameters
\(\theta_j\). The parent sets define the Hovorka-inspired graph:
\begin{align}
\Delta Q_1 &= f_{Q_1}(Q_1,Q_2,x_1,x_3,G_2) \, , &
\Delta Q_2 &= f_{Q_2}(Q_1,Q_2,x_1,x_2) \, , \\
\Delta S_1 &= f_{S_1}(S_1,u^{\mathrm{ins}}) \, , &
\Delta S_2 &= f_{S_2}(S_1,S_2) \, , \\
\Delta I &= f_I(S_2,I) \, , &
\Delta x_i &= f_{x_i}(I,x_i) \, , \quad i=1,2,3 \, , \\
\Delta G_1 &= f_{G_1}(G_1,u^{\mathrm{carb}}) \, , &
\Delta G_2 &= f_{G_2}(G_1,G_2) \, .
\end{align}

An LSTM encoder maps the observed history of glucose, insulin, and carbohydrate
inputs to the initial latent state. The graph-structured decoder then rolls the
state forward under future insulin and carbohydrate drivers. Glucose is read
out from the accessible glucose compartment using the Hovorka measurement
relation:
\begin{equation}
\hat{y}_t = \frac{Q_{1,t}}{V_G} \, ,
\qquad
V_G = \mathrm{softplus}(\omega_{V_G})+\epsilon \,  ,
\end{equation}
where $\omega_{V_G}$ is an unconstrained learnable scalar and the softplus
parameterization ensures that $V_G$ remains positive.

\section{Full Results}
\label{apx:full_results}

{\small
\setlength{\tabcolsep}{3pt}
\renewcommand{\arraystretch}{1.15}
\begin{longtable}{@{}lc*{7}{c}@{}}
\caption{
    Full \(\alpha\)-sweep results across model families. For IFM,
    \(\alpha=\lambda_{\mathrm{jac}}\); for baselines, \(\alpha\) is the
    H2NCM prediction--ranking tradeoff. Results are averaged across seeds, with
    sample standard deviations reported below each mean. Lower RMSE is better;
    higher intervention metrics are better.
    }\label{tab:alpha_sweep_selected}\\
\toprule
Model & $\alpha$ & RMSE$_{30}$ & RMSE$_{60}$ & RMSE$_{120}$ & $S_{\mathrm{ins}}$ & $S_{\mathrm{carb}}$ & $D_{\mathrm{strict}}$ & $R_{\mathrm{strict}}$ \\
\midrule
\endfirsthead
\toprule
Model & $\alpha$ & RMSE$_{30}$ & RMSE$_{60}$ & RMSE$_{120}$ & $S_{\mathrm{ins}}$ & $S_{\mathrm{carb}}$ & $D_{\mathrm{strict}}$ & $R_{\mathrm{strict}}$ \\
\midrule
\endhead
\bottomrule
\endlastfoot
IFM & 0.0 & \shortstack[c]{\textbf{8.76}\\{\scriptsize $\pm$ \textbf{0.55}}} & \shortstack[c]{\textbf{12.65}\\{\scriptsize $\pm$ \textbf{0.84}}} & \shortstack[c]{\textbf{18.30}\\{\scriptsize $\pm$ \textbf{1.11}}} & \shortstack[c]{\textbf{15.80}\\{\scriptsize $\pm$ \textbf{1.12}}} & \shortstack[c]{17.08\\{\scriptsize $\pm$ 2.97}} & \shortstack[c]{0.478\\{\scriptsize $\pm$ 0.001}} & \shortstack[c]{0.002\\{\scriptsize $\pm$ 0.002}} \\
IFM & 0.1 & \shortstack[c]{9.11\\{\scriptsize $\pm$ 0.45}} & \shortstack[c]{14.81\\{\scriptsize $\pm$ 0.81}} & \shortstack[c]{23.53\\{\scriptsize $\pm$ 0.99}} & \shortstack[c]{6.17\\{\scriptsize $\pm$ 0.56}} & \shortstack[c]{36.02\\{\scriptsize $\pm$ 2.73}} & \shortstack[c]{0.936\\{\scriptsize $\pm$ 0.004}} & \shortstack[c]{0.914\\{\scriptsize $\pm$ 0.007}} \\
IFM & 0.3 & \shortstack[c]{9.10\\{\scriptsize $\pm$ 0.90}} & \shortstack[c]{14.34\\{\scriptsize $\pm$ 0.79}} & \shortstack[c]{23.40\\{\scriptsize $\pm$ 1.12}} & \shortstack[c]{7.49\\{\scriptsize $\pm$ 0.27}} & \shortstack[c]{\textbf{39.34}\\{\scriptsize $\pm$ \textbf{4.19}}} & \shortstack[c]{0.937\\{\scriptsize $\pm$ 0.000}} & \shortstack[c]{0.917\\{\scriptsize $\pm$ 0.000}} \\
IFM & 0.5 & \shortstack[c]{9.31\\{\scriptsize $\pm$ 0.68}} & \shortstack[c]{14.71\\{\scriptsize $\pm$ 0.36}} & \shortstack[c]{24.05\\{\scriptsize $\pm$ 0.24}} & \shortstack[c]{7.35\\{\scriptsize $\pm$ 0.12}} & \shortstack[c]{35.72\\{\scriptsize $\pm$ 3.42}} & \shortstack[c]{0.938\\{\scriptsize $\pm$ 0.000}} & \shortstack[c]{0.917\\{\scriptsize $\pm$ 0.000}} \\
IFM & 0.7 & \shortstack[c]{9.01\\{\scriptsize $\pm$ 0.83}} & \shortstack[c]{14.40\\{\scriptsize $\pm$ 0.51}} & \shortstack[c]{23.91\\{\scriptsize $\pm$ 0.67}} & \shortstack[c]{7.92\\{\scriptsize $\pm$ 0.30}} & \shortstack[c]{35.23\\{\scriptsize $\pm$ 3.19}} & \shortstack[c]{\textbf{0.938}\\{\scriptsize $\pm$ \textbf{0.000}}} & \shortstack[c]{0.918\\{\scriptsize $\pm$ 0.000}} \\
IFM & 1.0 & \shortstack[c]{9.11\\{\scriptsize $\pm$ 0.59}} & \shortstack[c]{14.74\\{\scriptsize $\pm$ 0.78}} & \shortstack[c]{24.90\\{\scriptsize $\pm$ 1.75}} & \shortstack[c]{7.59\\{\scriptsize $\pm$ 0.44}} & \shortstack[c]{37.92\\{\scriptsize $\pm$ 3.44}} & \shortstack[c]{0.938\\{\scriptsize $\pm$ 0.000}} & \shortstack[c]{\textbf{0.918}\\{\scriptsize $\pm$ \textbf{0.000}}} \\
\midrule
Hovorka & 0.1 & \shortstack[c]{\textbf{8.24}\\{\scriptsize $\pm$ \textbf{1.29}}} & \shortstack[c]{13.26\\{\scriptsize $\pm$ 1.50}} & \shortstack[c]{\textbf{19.36}\\{\scriptsize $\pm$ \textbf{0.91}}} & \shortstack[c]{2.94\\{\scriptsize $\pm$ 1.59}} & \shortstack[c]{41.48\\{\scriptsize $\pm$ 8.87}} & \shortstack[c]{\textbf{0.762}\\{\scriptsize $\pm$ \textbf{0.018}}} & \shortstack[c]{\textbf{0.609}\\{\scriptsize $\pm$ \textbf{0.036}}} \\
Hovorka & 0.3 & \shortstack[c]{8.32\\{\scriptsize $\pm$ 0.66}} & \shortstack[c]{\textbf{13.21}\\{\scriptsize $\pm$ \textbf{0.96}}} & \shortstack[c]{20.52\\{\scriptsize $\pm$ 1.09}} & \shortstack[c]{1.30\\{\scriptsize $\pm$ 0.52}} & \shortstack[c]{50.14\\{\scriptsize $\pm$ 1.79}} & \shortstack[c]{0.742\\{\scriptsize $\pm$ 0.051}} & \shortstack[c]{0.569\\{\scriptsize $\pm$ 0.101}} \\
Hovorka & 0.5 & \shortstack[c]{8.50\\{\scriptsize $\pm$ 0.61}} & \shortstack[c]{13.42\\{\scriptsize $\pm$ 0.71}} & \shortstack[c]{20.96\\{\scriptsize $\pm$ 0.74}} & \shortstack[c]{1.18\\{\scriptsize $\pm$ 0.08}} & \shortstack[c]{51.39\\{\scriptsize $\pm$ 5.60}} & \shortstack[c]{0.743\\{\scriptsize $\pm$ 0.021}} & \shortstack[c]{0.570\\{\scriptsize $\pm$ 0.043}} \\
Hovorka & 0.7 & \shortstack[c]{8.92\\{\scriptsize $\pm$ 1.01}} & \shortstack[c]{14.02\\{\scriptsize $\pm$ 0.71}} & \shortstack[c]{21.38\\{\scriptsize $\pm$ 0.82}} & \shortstack[c]{1.07\\{\scriptsize $\pm$ 0.14}} & \shortstack[c]{51.39\\{\scriptsize $\pm$ 5.10}} & \shortstack[c]{0.747\\{\scriptsize $\pm$ 0.033}} & \shortstack[c]{0.578\\{\scriptsize $\pm$ 0.066}} \\
Hovorka & 1.0 & \shortstack[c]{55.49\\{\scriptsize $\pm$ 24.93}} & \shortstack[c]{165.34\\{\scriptsize $\pm$ 54.18}} & \shortstack[c]{2706.45\\{\scriptsize $\pm$ 2665.93}} & \shortstack[c]{\textbf{161.11}\\{\scriptsize $\pm$ \textbf{155.12}}} & \shortstack[c]{\textbf{364.80}\\{\scriptsize $\pm$ \textbf{405.33}}} & \shortstack[c]{0.701\\{\scriptsize $\pm$ 0.022}} & \shortstack[c]{0.366\\{\scriptsize $\pm$ 0.115}} \\
\midrule
BNODE & 0.1 & \shortstack[c]{\textbf{8.93}\\{\scriptsize $\pm$ \textbf{0.68}}} & \shortstack[c]{\textbf{13.17}\\{\scriptsize $\pm$ \textbf{1.12}}} & \shortstack[c]{\textbf{18.63}\\{\scriptsize $\pm$ \textbf{1.61}}} & \shortstack[c]{\textbf{1.67}\\{\scriptsize $\pm$ \textbf{0.36}}} & \shortstack[c]{23.56\\{\scriptsize $\pm$ 3.46}} & \shortstack[c]{0.883\\{\scriptsize $\pm$ 0.030}} & \shortstack[c]{0.765\\{\scriptsize $\pm$ 0.059}} \\
BNODE & 0.3 & \shortstack[c]{9.25\\{\scriptsize $\pm$ 0.95}} & \shortstack[c]{13.67\\{\scriptsize $\pm$ 1.22}} & \shortstack[c]{19.60\\{\scriptsize $\pm$ 1.21}} & \shortstack[c]{1.32\\{\scriptsize $\pm$ 0.25}} & \shortstack[c]{27.40\\{\scriptsize $\pm$ 4.39}} & \shortstack[c]{0.879\\{\scriptsize $\pm$ 0.029}} & \shortstack[c]{0.757\\{\scriptsize $\pm$ 0.058}} \\
BNODE & 0.5 & \shortstack[c]{9.31\\{\scriptsize $\pm$ 0.92}} & \shortstack[c]{13.87\\{\scriptsize $\pm$ 1.22}} & \shortstack[c]{19.73\\{\scriptsize $\pm$ 1.31}} & \shortstack[c]{1.17\\{\scriptsize $\pm$ 0.28}} & \shortstack[c]{30.82\\{\scriptsize $\pm$ 3.86}} & \shortstack[c]{0.874\\{\scriptsize $\pm$ 0.017}} & \shortstack[c]{0.748\\{\scriptsize $\pm$ 0.034}} \\
BNODE & 0.7 & \shortstack[c]{9.56\\{\scriptsize $\pm$ 0.91}} & \shortstack[c]{14.38\\{\scriptsize $\pm$ 1.28}} & \shortstack[c]{20.91\\{\scriptsize $\pm$ 1.82}} & \shortstack[c]{1.14\\{\scriptsize $\pm$ 0.22}} & \shortstack[c]{35.77\\{\scriptsize $\pm$ 12.96}} & \shortstack[c]{\textbf{0.941}\\{\scriptsize $\pm$ \textbf{0.031}}} & \shortstack[c]{\textbf{0.894}\\{\scriptsize $\pm$ \textbf{0.041}}} \\
BNODE & 1.0 & \shortstack[c]{187.88\\{\scriptsize $\pm$ 24.91}} & \shortstack[c]{256.48\\{\scriptsize $\pm$ 59.25}} & \shortstack[c]{1996.69\\{\scriptsize $\pm$ 1507.35}} & -- & \shortstack[c]{\textbf{63.57}\\{\scriptsize $\pm$ \textbf{41.46}}} & -- & -- \\
\midrule
LP-Reduced & 0.1 & \shortstack[c]{\textbf{10.44}\\{\scriptsize $\pm$ \textbf{1.65}}} & \shortstack[c]{\textbf{16.80}\\{\scriptsize $\pm$ \textbf{3.14}}} & \shortstack[c]{\textbf{25.50}\\{\scriptsize $\pm$ \textbf{4.23}}} & \shortstack[c]{0.39\\{\scriptsize $\pm$ 0.03}} & \shortstack[c]{24.64\\{\scriptsize $\pm$ 16.11}} & \shortstack[c]{0.560\\{\scriptsize $\pm$ 0.094}} & \shortstack[c]{0.324\\{\scriptsize $\pm$ 0.039}} \\
LP-Reduced & 0.3 & \shortstack[c]{11.33\\{\scriptsize $\pm$ 1.96}} & \shortstack[c]{18.79\\{\scriptsize $\pm$ 4.41}} & \shortstack[c]{28.77\\{\scriptsize $\pm$ 6.70}} & \shortstack[c]{0.61\\{\scriptsize $\pm$ 0.09}} & \shortstack[c]{18.71\\{\scriptsize $\pm$ 18.18}} & \shortstack[c]{0.632\\{\scriptsize $\pm$ 0.042}} & \shortstack[c]{0.441\\{\scriptsize $\pm$ 0.046}} \\
LP-Reduced & 0.5 & \shortstack[c]{13.12\\{\scriptsize $\pm$ 2.15}} & \shortstack[c]{21.57\\{\scriptsize $\pm$ 3.48}} & \shortstack[c]{32.10\\{\scriptsize $\pm$ 5.47}} & \shortstack[c]{0.62\\{\scriptsize $\pm$ 0.07}} & \shortstack[c]{19.58\\{\scriptsize $\pm$ 20.10}} & \shortstack[c]{0.622\\{\scriptsize $\pm$ 0.069}} & \shortstack[c]{0.462\\{\scriptsize $\pm$ 0.071}} \\
LP-Reduced & 0.7 & \shortstack[c]{18.73\\{\scriptsize $\pm$ 12.51}} & \shortstack[c]{24.54\\{\scriptsize $\pm$ 10.88}} & \shortstack[c]{33.28\\{\scriptsize $\pm$ 9.53}} & \shortstack[c]{0.67\\{\scriptsize $\pm$ 0.05}} & \shortstack[c]{\textbf{24.75}\\{\scriptsize $\pm$ \textbf{16.85}}} & \shortstack[c]{\textbf{0.667}\\{\scriptsize $\pm$ \textbf{0.036}}} & \shortstack[c]{0.497\\{\scriptsize $\pm$ 0.111}} \\
LP-Reduced & 1.0 & \shortstack[c]{112.80\\{\scriptsize $\pm$ 112.90}} & \shortstack[c]{108.00\\{\scriptsize $\pm$ 65.29}} & \shortstack[c]{113.62\\{\scriptsize $\pm$ 38.18}} & \shortstack[c]{\textbf{0.86}\\{\scriptsize $\pm$ \textbf{0.49}}} & \shortstack[c]{8.65\\{\scriptsize $\pm$ 7.56}} & \shortstack[c]{0.666\\{\scriptsize $\pm$ 0.125}} & \shortstack[c]{\textbf{0.522}\\{\scriptsize $\pm$ \textbf{0.200}}} \\
\midrule
LSTM & 0.1 & \shortstack[c]{5.16\\{\scriptsize $\pm$ 0.42}} & \shortstack[c]{\textbf{7.77}\\{\scriptsize $\pm$ \textbf{0.70}}} & \shortstack[c]{\textbf{11.30}\\{\scriptsize $\pm$ \textbf{0.70}}} & \shortstack[c]{5.09\\{\scriptsize $\pm$ 0.21}} & \shortstack[c]{14.36\\{\scriptsize $\pm$ 1.69}} & \shortstack[c]{0.837\\{\scriptsize $\pm$ 0.011}} & \shortstack[c]{0.677\\{\scriptsize $\pm$ 0.021}} \\
LSTM & 0.3 & \shortstack[c]{\textbf{4.89}\\{\scriptsize $\pm$ \textbf{0.17}}} & \shortstack[c]{7.84\\{\scriptsize $\pm$ 0.47}} & \shortstack[c]{11.90\\{\scriptsize $\pm$ 0.60}} & \shortstack[c]{4.02\\{\scriptsize $\pm$ 0.38}} & \shortstack[c]{14.49\\{\scriptsize $\pm$ 0.69}} & \shortstack[c]{0.859\\{\scriptsize $\pm$ 0.029}} & \shortstack[c]{0.722\\{\scriptsize $\pm$ 0.060}} \\
LSTM & 0.5 & \shortstack[c]{5.06\\{\scriptsize $\pm$ 0.24}} & \shortstack[c]{8.01\\{\scriptsize $\pm$ 0.54}} & \shortstack[c]{12.39\\{\scriptsize $\pm$ 0.66}} & \shortstack[c]{3.52\\{\scriptsize $\pm$ 0.25}} & \shortstack[c]{16.32\\{\scriptsize $\pm$ 1.75}} & \shortstack[c]{0.873\\{\scriptsize $\pm$ 0.048}} & \shortstack[c]{0.754\\{\scriptsize $\pm$ 0.091}} \\
LSTM & 0.7 & \shortstack[c]{5.01\\{\scriptsize $\pm$ 0.51}} & \shortstack[c]{8.29\\{\scriptsize $\pm$ 0.78}} & \shortstack[c]{13.11\\{\scriptsize $\pm$ 0.61}} & \shortstack[c]{3.29\\{\scriptsize $\pm$ 0.85}} & \shortstack[c]{\textbf{19.35}\\{\scriptsize $\pm$ \textbf{3.42}}} & \shortstack[c]{0.899\\{\scriptsize $\pm$ 0.050}} & \shortstack[c]{0.803\\{\scriptsize $\pm$ 0.101}} \\
LSTM & 1.0 & \shortstack[c]{61.55\\{\scriptsize $\pm$ 4.08}} & \shortstack[c]{61.42\\{\scriptsize $\pm$ 3.82}} & \shortstack[c]{62.57\\{\scriptsize $\pm$ 3.72}} & \shortstack[c]{\textbf{5.33}\\{\scriptsize $\pm$ \textbf{1.09}}} & \shortstack[c]{11.28\\{\scriptsize $\pm$ 2.15}} & \shortstack[c]{\textbf{0.958}\\{\scriptsize $\pm$ \textbf{0.000}}} & \shortstack[c]{\textbf{0.917}\\{\scriptsize $\pm$ \textbf{0.000}}} \\
\midrule
S4D & 0.1 & \shortstack[c]{\textbf{13.00}\\{\scriptsize $\pm$ \textbf{0.46}}} & \shortstack[c]{\textbf{13.81}\\{\scriptsize $\pm$ \textbf{0.54}}} & \shortstack[c]{\textbf{14.81}\\{\scriptsize $\pm$ \textbf{0.47}}} & \shortstack[c]{\textbf{2.84}\\{\scriptsize $\pm$ \textbf{0.21}}} & \shortstack[c]{16.09\\{\scriptsize $\pm$ 1.57}} & \shortstack[c]{0.763\\{\scriptsize $\pm$ 0.015}} & \shortstack[c]{0.533\\{\scriptsize $\pm$ 0.031}} \\
S4D & 0.3 & \shortstack[c]{13.70\\{\scriptsize $\pm$ 0.46}} & \shortstack[c]{14.63\\{\scriptsize $\pm$ 0.64}} & \shortstack[c]{15.67\\{\scriptsize $\pm$ 0.50}} & \shortstack[c]{2.46\\{\scriptsize $\pm$ 0.15}} & \shortstack[c]{18.03\\{\scriptsize $\pm$ 0.89}} & \shortstack[c]{0.786\\{\scriptsize $\pm$ 0.017}} & \shortstack[c]{0.574\\{\scriptsize $\pm$ 0.034}} \\
S4D & 0.5 & \shortstack[c]{14.56\\{\scriptsize $\pm$ 0.36}} & \shortstack[c]{15.52\\{\scriptsize $\pm$ 0.63}} & \shortstack[c]{16.57\\{\scriptsize $\pm$ 0.59}} & \shortstack[c]{2.16\\{\scriptsize $\pm$ 0.10}} & \shortstack[c]{\textbf{19.41}\\{\scriptsize $\pm$ \textbf{1.51}}} & \shortstack[c]{0.771\\{\scriptsize $\pm$ 0.013}} & \shortstack[c]{0.543\\{\scriptsize $\pm$ 0.026}} \\
S4D & 0.7 & \shortstack[c]{15.46\\{\scriptsize $\pm$ 0.60}} & \shortstack[c]{16.43\\{\scriptsize $\pm$ 0.80}} & \shortstack[c]{17.52\\{\scriptsize $\pm$ 0.71}} & \shortstack[c]{1.97\\{\scriptsize $\pm$ 0.12}} & \shortstack[c]{18.89\\{\scriptsize $\pm$ 0.75}} & \shortstack[c]{0.776\\{\scriptsize $\pm$ 0.012}} & \shortstack[c]{0.553\\{\scriptsize $\pm$ 0.024}} \\
S4D & 1.0 & \shortstack[c]{61.33\\{\scriptsize $\pm$ 3.79}} & \shortstack[c]{60.47\\{\scriptsize $\pm$ 3.85}} & \shortstack[c]{60.25\\{\scriptsize $\pm$ 3.85}} & \shortstack[c]{1.03\\{\scriptsize $\pm$ 0.18}} & \shortstack[c]{1.19\\{\scriptsize $\pm$ 0.09}} & \shortstack[c]{\textbf{0.874}\\{\scriptsize $\pm$ \textbf{0.017}}} & \shortstack[c]{\textbf{0.753}\\{\scriptsize $\pm$ \textbf{0.032}}} \\
\midrule
TCN & 0.1 & \shortstack[c]{\textbf{8.13}\\{\scriptsize $\pm$ \textbf{0.33}}} & \shortstack[c]{\textbf{9.81}\\{\scriptsize $\pm$ \textbf{0.46}}} & \shortstack[c]{\textbf{12.06}\\{\scriptsize $\pm$ \textbf{0.50}}} & \shortstack[c]{\textbf{9.28}\\{\scriptsize $\pm$ \textbf{0.53}}} & \shortstack[c]{13.90\\{\scriptsize $\pm$ 5.21}} & \shortstack[c]{0.829\\{\scriptsize $\pm$ 0.023}} & \shortstack[c]{0.660\\{\scriptsize $\pm$ 0.047}} \\
TCN & 0.3 & \shortstack[c]{8.76\\{\scriptsize $\pm$ 0.37}} & \shortstack[c]{10.75\\{\scriptsize $\pm$ 0.53}} & \shortstack[c]{13.45\\{\scriptsize $\pm$ 0.79}} & \shortstack[c]{8.72\\{\scriptsize $\pm$ 0.29}} & \shortstack[c]{15.62\\{\scriptsize $\pm$ 2.42}} & \shortstack[c]{0.815\\{\scriptsize $\pm$ 0.013}} & \shortstack[c]{0.630\\{\scriptsize $\pm$ 0.025}} \\
TCN & 0.5 & \shortstack[c]{9.46\\{\scriptsize $\pm$ 0.52}} & \shortstack[c]{11.38\\{\scriptsize $\pm$ 0.69}} & \shortstack[c]{14.31\\{\scriptsize $\pm$ 0.59}} & \shortstack[c]{8.48\\{\scriptsize $\pm$ 0.29}} & \shortstack[c]{16.93\\{\scriptsize $\pm$ 5.49}} & \shortstack[c]{0.804\\{\scriptsize $\pm$ 0.013}} & \shortstack[c]{0.608\\{\scriptsize $\pm$ 0.026}} \\
TCN & 0.7 & \shortstack[c]{9.81\\{\scriptsize $\pm$ 0.57}} & \shortstack[c]{12.10\\{\scriptsize $\pm$ 0.69}} & \shortstack[c]{15.40\\{\scriptsize $\pm$ 0.96}} & \shortstack[c]{8.31\\{\scriptsize $\pm$ 0.62}} & \shortstack[c]{\textbf{29.58}\\{\scriptsize $\pm$ \textbf{9.42}}} & \shortstack[c]{0.795\\{\scriptsize $\pm$ 0.002}} & \shortstack[c]{0.589\\{\scriptsize $\pm$ 0.004}} \\
TCN & 1.0 & \shortstack[c]{62.31\\{\scriptsize $\pm$ 4.01}} & \shortstack[c]{61.78\\{\scriptsize $\pm$ 4.09}} & \shortstack[c]{62.01\\{\scriptsize $\pm$ 3.97}} & \shortstack[c]{3.95\\{\scriptsize $\pm$ 1.05}} & \shortstack[c]{7.00\\{\scriptsize $\pm$ 0.65}} & \shortstack[c]{\textbf{1.000}\\{\scriptsize $\pm$ \textbf{0.000}}} & \shortstack[c]{\textbf{1.000}\\{\scriptsize $\pm$ \textbf{0.000}}} \\
\end{longtable}
}

{\small
\setlength{\tabcolsep}{3pt}
\renewcommand{\arraystretch}{1.15}
\begin{longtable}{@{}c*{8}{c}@{}}
\caption{
Fidelity subsampling-ratio sweep for IFM. Each
entry reports mean \(\pm\) sample standard deviation across seeds. Lower RMSE is
better; higher intervention metrics are better.
}\label{tab:fidelity_subsample_ratio_sweep_selected}\\
\toprule
$\rho$ & RMSE$_{30}$ & RMSE$_{60}$ & RMSE$_{120}$ & $S_{\mathrm{ins}}$ & $S_{\mathrm{carb}}$ & $D_{\mathrm{strict}}$ & $R_{\mathrm{strict}}$ \\
\midrule
\endfirsthead
\toprule
$\rho_{\mathrm{fid}}$ & RMSE$_{30}$ & RMSE$_{60}$ & RMSE$_{120}$ & $S_{\mathrm{ins}}$ & $S_{\mathrm{carb}}$ & $D_{\mathrm{strict}}$ & $R_{\mathrm{strict}}$ \\
\midrule
\endhead
\bottomrule
\endlastfoot
0.1 & \shortstack[c]{8.93\\{\scriptsize $\pm$ 0.41}} & \shortstack[c]{14.26\\{\scriptsize $\pm$ 1.15}} & \shortstack[c]{23.32\\{\scriptsize $\pm$ 1.79}} & \shortstack[c]{\textbf{7.26}\\{\scriptsize $\pm$ \textbf{1.24}}} & \shortstack[c]{39.89\\{\scriptsize $\pm$ 3.33}} & \shortstack[c]{\textbf{0.936}\\{\scriptsize $\pm$ \textbf{0.005}}} & \shortstack[c]{\textbf{0.915}\\{\scriptsize $\pm$ \textbf{0.010}}} \\
0.3 & \shortstack[c]{8.91\\{\scriptsize $\pm$ 0.35}} & \shortstack[c]{13.78\\{\scriptsize $\pm$ 0.93}} & \shortstack[c]{22.43\\{\scriptsize $\pm$ 1.17}} & \shortstack[c]{6.78\\{\scriptsize $\pm$ 0.64}} & \shortstack[c]{38.90\\{\scriptsize $\pm$ 4.23}} & \shortstack[c]{0.935\\{\scriptsize $\pm$ 0.003}} & \shortstack[c]{0.913\\{\scriptsize $\pm$ 0.006}} \\
0.5 & \shortstack[c]{9.92\\{\scriptsize $\pm$ 1.22}} & \shortstack[c]{14.22\\{\scriptsize $\pm$ 1.38}} & \shortstack[c]{21.98\\{\scriptsize $\pm$ 1.30}} & \shortstack[c]{7.07\\{\scriptsize $\pm$ 0.87}} & \shortstack[c]{40.17\\{\scriptsize $\pm$ 1.92}} & \shortstack[c]{0.933\\{\scriptsize $\pm$ 0.006}} & \shortstack[c]{0.909\\{\scriptsize $\pm$ 0.011}} \\
0.7 & \shortstack[c]{8.41\\{\scriptsize $\pm$ 1.13}} & \shortstack[c]{\textbf{13.09}\\{\scriptsize $\pm$ \textbf{1.31}}} & \shortstack[c]{\textbf{21.16}\\{\scriptsize $\pm$ \textbf{1.22}}} & \shortstack[c]{6.63\\{\scriptsize $\pm$ 0.65}} & \shortstack[c]{40.39\\{\scriptsize $\pm$ 3.77}} & \shortstack[c]{0.928\\{\scriptsize $\pm$ 0.003}} & \shortstack[c]{0.898\\{\scriptsize $\pm$ 0.006}} \\
1 & \shortstack[c]{\textbf{7.69}\\{\scriptsize $\pm$ \textbf{0.92}}} & \shortstack[c]{13.24\\{\scriptsize $\pm$ 0.97}} & \shortstack[c]{21.91\\{\scriptsize $\pm$ 0.68}} & \shortstack[c]{7.23\\{\scriptsize $\pm$ 0.55}} & \shortstack[c]{\textbf{42.05}\\{\scriptsize $\pm$ \textbf{3.29}}} & \shortstack[c]{0.933\\{\scriptsize $\pm$ 0.004}} & \shortstack[c]{0.908\\{\scriptsize $\pm$ 0.007}} \\
\end{longtable}
}

{\small
\setlength{\tabcolsep}{3pt}
\renewcommand{\arraystretch}{1.15}
\begin{longtable}{@{}c*{7}{c}@{}}
\caption{
Fidelity-loss weight sweep for IFM. Each entry reports
mean \(\pm\) sample standard deviation across seeds. Lower RMSE is better;
higher intervention metrics are better.
}\label{tab:lambda_fidelity_sweep_selected}\\
\toprule
$\lambda_{\mathrm{fid}}$ & RMSE$_{30}$ & RMSE$_{60}$ & RMSE$_{120}$ & $S_{\mathrm{ins}}$ & $S_{\mathrm{carb}}$ & $D_{\mathrm{strict}}$ & $R_{\mathrm{strict}}$ \\
\midrule
\endfirsthead
\toprule
$\lambda_{\mathrm{fid}}$ & RMSE$_{30}$ & RMSE$_{60}$ & RMSE$_{120}$ & $S_{\mathrm{ins}}$ & $S_{\mathrm{carb}}$ & $D_{\mathrm{strict}}$ & $R_{\mathrm{strict}}$ \\
\midrule
\endhead
\bottomrule
\endlastfoot
0 & \shortstack[c]{11.66\\{\scriptsize $\pm$ 0.92}} & \shortstack[c]{25.16\\{\scriptsize $\pm$ 0.71}} & \shortstack[c]{52.10\\{\scriptsize $\pm$ 2.26}} & \shortstack[c]{\textbf{29.82}\\{\scriptsize $\pm$ \textbf{3.10}}} & \shortstack[c]{\textbf{55.37}\\{\scriptsize $\pm$ \textbf{4.33}}} & \shortstack[c]{0.935\\{\scriptsize $\pm$ 0.000}} & \shortstack[c]{0.913\\{\scriptsize $\pm$ 0.000}} \\
0.1 & \shortstack[c]{9.62\\{\scriptsize $\pm$ 0.72}} & \shortstack[c]{15.48\\{\scriptsize $\pm$ 0.48}} & \shortstack[c]{26.31\\{\scriptsize $\pm$ 0.71}} & \shortstack[c]{11.59\\{\scriptsize $\pm$ 0.97}} & \shortstack[c]{42.26\\{\scriptsize $\pm$ 6.72}} & \shortstack[c]{0.934\\{\scriptsize $\pm$ 0.002}} & \shortstack[c]{0.910\\{\scriptsize $\pm$ 0.003}} \\
0.3 & \shortstack[c]{\textbf{8.60}\\{\scriptsize $\pm$ \textbf{0.30}}} & \shortstack[c]{15.01\\{\scriptsize $\pm$ 0.75}} & \shortstack[c]{25.61\\{\scriptsize $\pm$ 1.42}} & \shortstack[c]{10.10\\{\scriptsize $\pm$ 0.39}} & \shortstack[c]{40.09\\{\scriptsize $\pm$ 7.68}} & \shortstack[c]{0.936\\{\scriptsize $\pm$ 0.003}} & \shortstack[c]{0.914\\{\scriptsize $\pm$ 0.007}} \\
0.5 & \shortstack[c]{9.31\\{\scriptsize $\pm$ 0.78}} & \shortstack[c]{15.34\\{\scriptsize $\pm$ 1.17}} & \shortstack[c]{25.45\\{\scriptsize $\pm$ 1.34}} & \shortstack[c]{8.34\\{\scriptsize $\pm$ 1.04}} & \shortstack[c]{36.53\\{\scriptsize $\pm$ 4.08}} & \shortstack[c]{0.935\\{\scriptsize $\pm$ 0.002}} & \shortstack[c]{0.912\\{\scriptsize $\pm$ 0.004}} \\
0.7 & \shortstack[c]{9.27\\{\scriptsize $\pm$ 0.93}} & \shortstack[c]{14.59\\{\scriptsize $\pm$ 0.85}} & \shortstack[c]{24.37\\{\scriptsize $\pm$ 1.10}} & \shortstack[c]{7.88\\{\scriptsize $\pm$ 0.48}} & \shortstack[c]{37.34\\{\scriptsize $\pm$ 5.46}} & \shortstack[c]{0.938\\{\scriptsize $\pm$ 0.000}} & \shortstack[c]{0.917\\{\scriptsize $\pm$ 0.000}} \\
1 & \shortstack[c]{9.24\\{\scriptsize $\pm$ 0.54}} & \shortstack[c]{\textbf{14.24}\\{\scriptsize $\pm$ \textbf{0.38}}} & \shortstack[c]{\textbf{23.71}\\{\scriptsize $\pm$ \textbf{0.62}}} & \shortstack[c]{7.25\\{\scriptsize $\pm$ 0.17}} & \shortstack[c]{36.22\\{\scriptsize $\pm$ 4.57}} & \shortstack[c]{\textbf{0.938}\\{\scriptsize $\pm$ \textbf{0.000}}} & \shortstack[c]{\textbf{0.918}\\{\scriptsize $\pm$ \textbf{0.000}}} \\
\end{longtable}
}

\section{Training Hyperparameters}
\label{apx:training_details}

\begin{longtable}{llcc}
\caption{Hyperparameter grid search spaces and selected optimal values for each model. We constrained the grid search to the first 20 combinations.}
\label{tab:hparam_search_optimal} \\
\toprule
\textbf{Model} & \textbf{Hyperparameter} & \textbf{Search Space} & \textbf{Selected} \\
\midrule
\endfirsthead

\toprule
\textbf{Model} & \textbf{Hyperparameter} & \textbf{Search Space} & \textbf{Selected} \\
\midrule
\endhead

\multirow{4}{*}{BNODE} & Latent size & $\{32, 64, 128\}$ & 32 \\
 & MLP size & $\{32, 64, 128\}$ & 128 \\
 & Num hidden layer & $\{1, 2, 3\}$ & 3 \\
 & Number of layers & $\{1, 2\}$ & 1 \\

\midrule

\multirow{5}{*}{IFM} & Flow hidden dimension & $\{32, 64, 128\}$ & 32 \\
 & Model dimension~$(d_f)$ & $\{32, 64\}$ & 32 \\
 & Number of encoder layers & $\{1, 2, 3\}$ & 3 \\
 & Number of flow layers & $\{1, 2, 3\}$ & 1 \\

\midrule

\multirow{4}{*}{Hovorka} & Encoder size & $\{32, 64, 128\}$ & 32 \\
 & Lstm num layers & $\{1, 2\}$ & 2 \\
 & MLP size & $\{32, 64, 128\}$ & 128 \\
 & Number of hidden layers & $\{1, 2, 3\}$ & 3 \\

\midrule

\multirow{4}{*}{LP-Reduced} & Latent size & $\{32, 64, 128\}$ & 64 \\
 & MLP size & $\{32, 64, 128\}$ & 32 \\
 & Num hidden layer & $\{1, 2, 3\}$ & 1 \\
 & Number of layers & $\{1, 2\}$ & 2 \\

\midrule

\multirow{2}{*}{LSTM} & Latent size & $\{32, 64, 128\}$ & 128 \\
 & Number of layers & $\{1, 2, 3\}$ & 3 \\

\midrule

\multirow{3}{*}{S4D} & D model & $\{32, 64, 128\}$ & 64 \\
 & D state & $\{32, 64, 128\}$ & 32 \\
 & Number of layers & $\{1, 2\}$ & 2 \\

\midrule

\multirow{3}{*}{TCN} & Channel width & $\{32, 64, 128\}$ & 64 \\
 & Kernel size & $\{2, 3, 5\}$ & 5 \\
 & Number of levels & $\{2, 3, 4\}$ & 4 \\
\bottomrule
\end{longtable}

\begin{table}[H]
\centering
\caption{Fixed hyperparameters used for all experiments.}
\label{tab:fixed_hyperparameters}
\begin{tabular}{lc}
\toprule
\textbf{Hyperparameter} & \textbf{Value} \\
\midrule
Weight decay & $1 \times 10^{-4}$ \\
Learning rate & $1 \times 10^{-3}$ \\
Batch size & 512 \\
Optimizer & AdamW~\citep{Loshchilov2017DecoupledWD} \\
Prediction horizon & 24 \\
Window size & 2016 \\
Flow steps $K_f$ & $20$ \\
Number of Monte Carlo Trials & $4$ \\
$g_{\min},g_{\max}$ & $40,400$ mg/dL \\
Skew parameter $\kappa$ & $1.0$ \\
$\alpha_{\mathrm{ins}}$ & $1.0$ units of insulin \\
$\alpha_{\mathrm{carb}}$ & $25.0$ grams\\
Insulin sensitivity bounds & $[-60,-30]$ mg/dL \\
Carbohydrate sensitivity bounds & $[5,80]$ mg/dL \\
Number of time frequencies $K$ & 8 \\

\bottomrule
\end{tabular}
\end{table}

\section{Ablation on PK Smoother and Physiological Sensitivity Bounds}
\label{apx:ablation}

\subsection{Effect of physiological driver smoothing.}
\label{apx:pk_smooth}
Removing the PK smoother and feeding raw drivers directly degrades both forecasting accuracy and intervention validity. Without smoothing, RMSE increases to \(11.95/21.45/34.74\) mg/dL at \(30/60/120\) minutes, compared with \(9.11/14.74/24.90\) with smoothing. The intervention metrics degrade more sharply: insulin sensitivity drops from \(7.59\) to \(2.18\), carbohydrate sensitivity from \(37.92\) to \(0.68\), directional consistency from \(0.938\) to \(0.706\), and ranking consistency from \(0.918\) to \(0.419\). This indicates that raw impulse inputs do not provide a physiologically aligned intervention pathway; smoothing is needed to expose delayed insulin and carbohydrate effects to the velocity network in a form that supports accurate rollout prediction and meaningful prospective intervention response.

\subsection{Physiological Sensitivity Bounds.}
\label{apx:local_bounds}
We sweep one channel's Jacobian bounds while holding the other fixed: carbohydrate bounds are fixed at \([5,80]\) during the insulin sweep, and insulin bounds are fixed at \([-60,-30]\) during the carbohydrate sweep. The carbohydrate sweep shows that increasing the carbohydrate upper bound mainly increases carbohydrate sensitivity, while strict direction and ranking remain stable. The insulin sweep is more sensitive: making the insulin interval more permissive, especially by moving the upper bound closer to zero, reduces insulin sensitivity substantially and weakens ranking and directional consistency. Overall, the bounds tune response magnitude, while their signs enforce the expected intervention direction. However, overly loose insulin bounds can move the constraint closer to the lazy solution \(E_{\mathrm{ins}}\approx0\), allowing the model to learn a much weaker glucose-lowering response. 

\begin{figure}[h]
    \centering
    \includegraphics[width=0.8\textwidth]{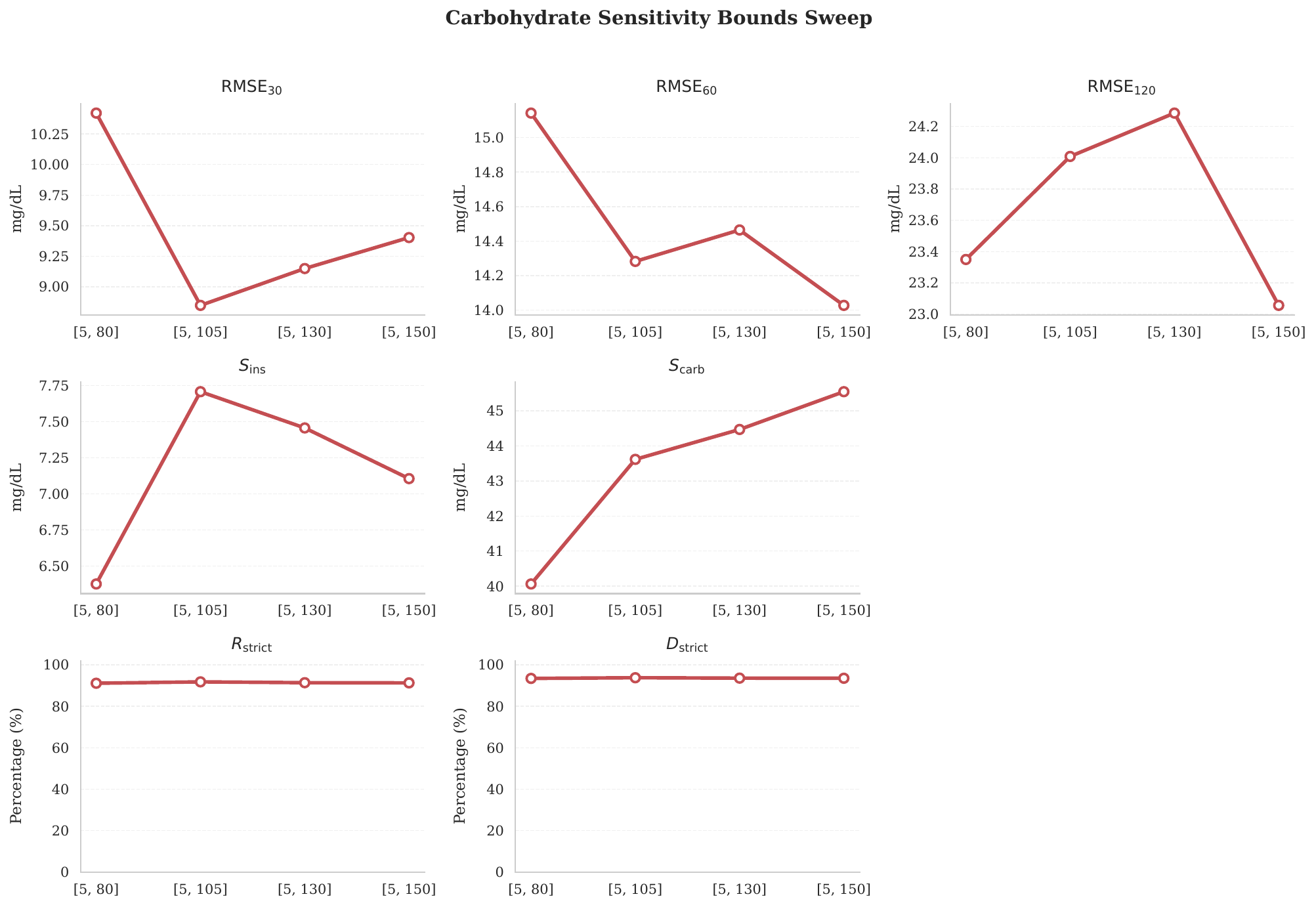}
    \caption{
    Carbohydrate-bound sweep with insulin bounds fixed at \([-60,-30]\). Increasing the carbohydrate upper bound mainly increases carbohydrate sensitivity, while strict direction and ranking remain stable across the tested intervals.
    }
    \label{fig:carb_bounds_sweep}
\end{figure}

\begin{figure}[h]
    \centering
    \includegraphics[width=0.8\textwidth]{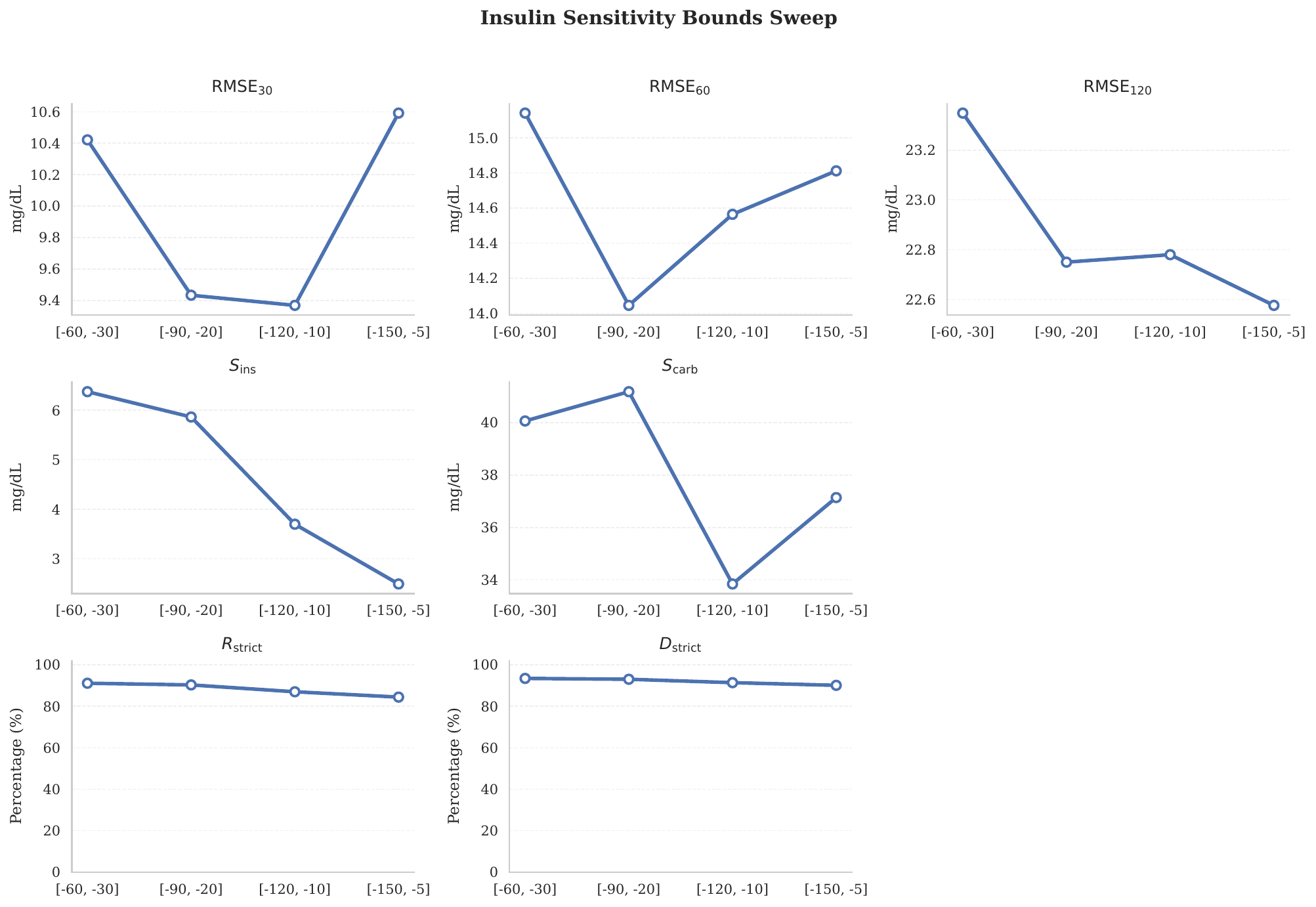}
    \caption{
    Insulin-bound sweep with carbohydrate bounds fixed at \([5,80]\). Looser insulin bounds reduce insulin sensitivity and weaken both ranking and directional consistency.
    }
    \label{fig:insulin_bounds_sweep}
\end{figure}

\subsection{Role of the Skew Parameter}
\label{apx:skew_param}

The skew parameter \(\kappa\) affects IFM through both the latent CFM target and
the physical-unit scaling of the Jacobian regularizer. In the latent
transform of Eq.~\eqref{eq:encode}, changing \(\kappa\) changes \(y\), and
therefore changes the encoded ground-truth latent target \(z_t^\star\) for the
same physical glucose value. Since the CFM target velocity is
\(v_t^\star=z_t^\star-z_{t,0}\), varying \(\kappa\) changes the latent transport
target learned by \(v_\theta\).

The same \(\kappa\) also appears in the decoder slope in Eq.~\eqref{eq:slope},
which is used in Eq.~\eqref{eq:sensitivity} to convert each \(J_c^{(b,t)}\) into the physical-unit sensitivity
\(E_c^{(b,t)}\). If the decoder slope becomes smaller in the relevant glucose range, the same
physiological sensitivity bounds \([\ell_c,u_c]\) require a larger-magnitude \(J_c^{(b,t)}\). In other words, the velocity field must
respond more strongly to the corresponding smoothed effective driver
\(d_t^{(b,c)}\) to produce a physical-unit sensitivity \(E_c^{(b,t)}\) inside the admissible bounds.

\begin{figure*}[h]
    \centering
    \includegraphics[width=1\textwidth]{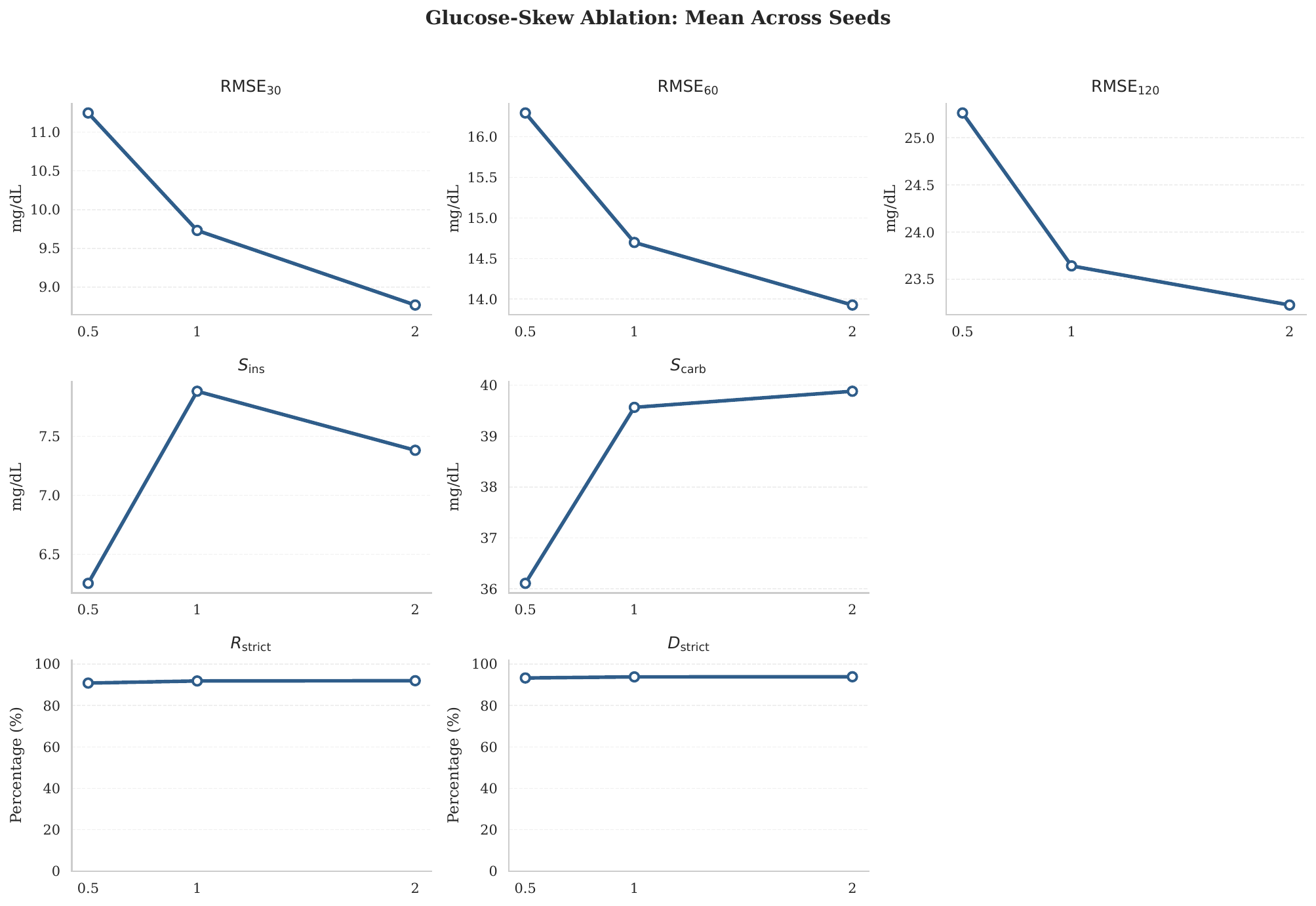}
    \caption{
    Effect of the skew parameter \(\kappa\). Increasing \(\kappa\) changes the
    latent glucose encoding and the decoder-slope scaling of the Jacobian
    regularizer, leading to lower observed-driver RMSE and higher intervention
    sensitivity magnitudes. Strict directional and ranking consistency remain
    stable across the sweep.
    }
    \label{fig:glucose_skew_ablation}
\end{figure*}

Figure~\ref{fig:glucose_skew_ablation} shows this effect empirically: changing
\(\kappa\) shifts the intervention sensitivity magnitudes, while directional
and ranking consistency remain stable.

\subsection{Existing Assets}
\label{apx:asset_licenses}

We use the public simulated glucose-management dataset
\citep{fathkouhi2026stationarity}, released under CC BY 4.0. We do not use the
UVA/Padova simulator directly.

\section{Qualitative Results}
\label{apx:qualitative}
\begin{figure}[h]
    \centering
    \includegraphics[width=0.95\textwidth]{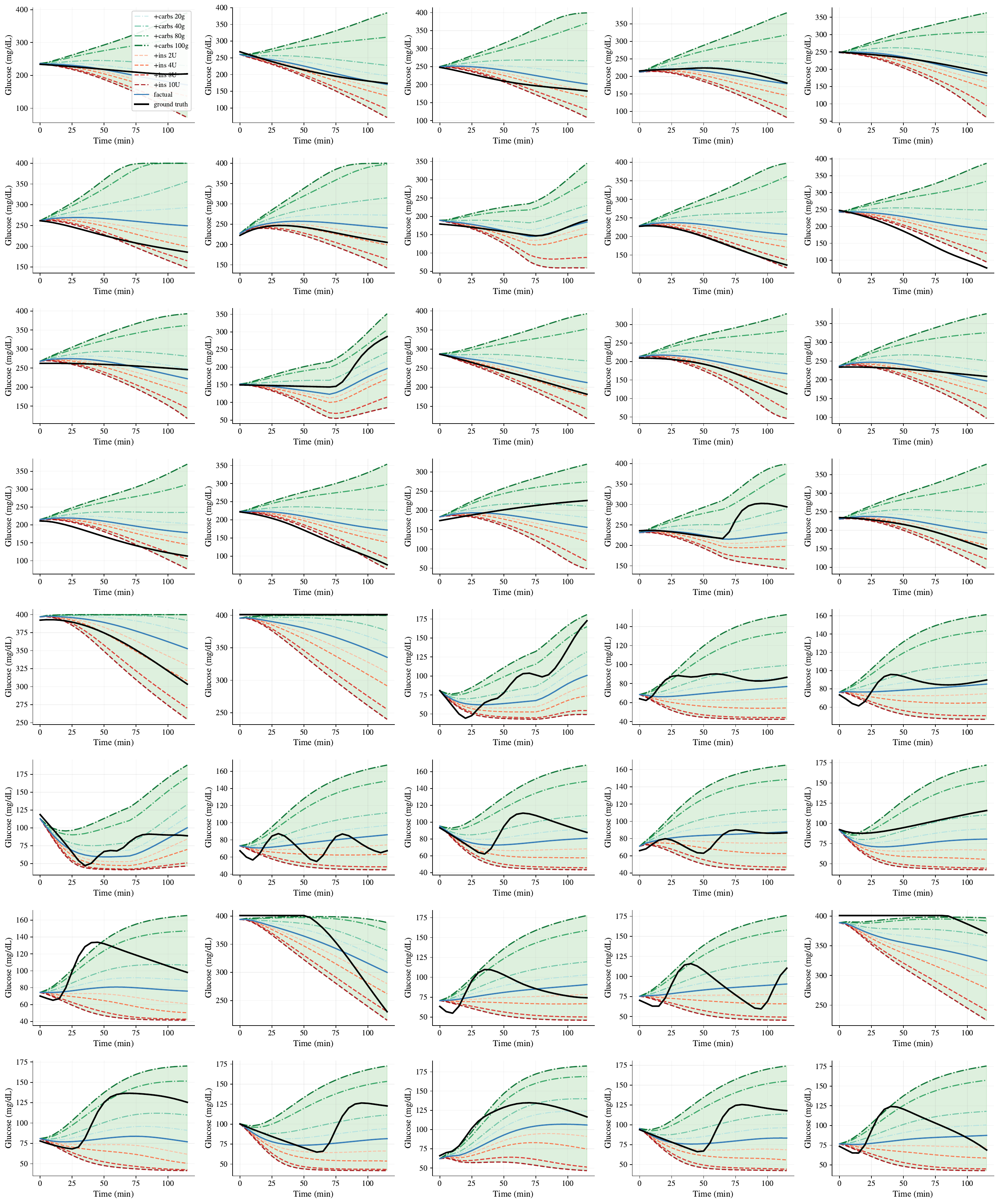}
    \caption{
    Example 24-step (2-hour) interventional forecasts from IFM under different planned insulin and carbohydrate perturbations. Insulin plans use doses of \(2\), \(4\), \(8\), and \(10\) units, while carbohydrate plans use \(20\), \(40\), \(80\), and \(100\) grams. Relative to the factual-driver forecast, increasing insulin produces smooth downward glucose responses, whereas increasing carbohydrate produces smooth upward responses. Near the decoder bounds, approximately \(40\) and \(400\) mg/dL, the trajectories saturate rather than extrapolating outside the physiological glucose range.
    }
    \label{fig:ifm_intervention_trajectories}
\end{figure}

\clearpage
\newpage
\section*{NeurIPS Paper Checklist}

\begin{enumerate}

\item {\bf Claims}
    \item[] Question: Do the main claims made in the abstract and introduction accurately reflect the paper's contributions and scope?
    \item[] Answer: \answerYes{}
    \item[] Justification:  The abstract and introduction clearly present IFM as a prospective dose-response forecasting method and state its scope as velocity-field response regularization.

\item {\bf Limitations}
    \item[] Question: Does the paper discuss the limitations of the work performed by the authors?
    \item[] Answer: \answerYes{}
    \item[] Justification: The paper discusses limitations including use of simulated data without real-world validation due to missing or noisy clinical carbohydrate records, and the rollout-based fidelity loss.

\item {\bf Theory assumptions and proofs}
    \item[] Question: For each theoretical result, does the paper provide the full set of assumptions and a complete (and correct) proof?
    \item[] Answer: \answerNA{}.
    \item[] Justification: No formal theorems are presented; the relevant mathematical steps are given directly in the methodology section.

    \item {\bf Experimental result reproducibility}
    \item[] Question: Does the paper fully disclose all the information needed to reproduce the main experimental results of the paper to the extent that it affects the main claims and/or conclusions of the paper (regardless of whether the code and data are provided or not)?
    \item[] Answer: \answerYes{}
    \item[] Justification: The paper provides the methodological details, experimental setup, hyperparameter search space, selected configurations, and evaluation metrics needed to reproduce the main results.

\item {\bf Open access to data and code}
    \item[] Question: Does the paper provide open access to the data and code, with sufficient instructions to faithfully reproduce the main experimental results, as described in supplemental material?
    \item[] Answer: \answerYes{}
    \item[] Justification: The dataset is publicly available, and the code with reproduction instructions is included in the supplemental material.

\item {\bf Experimental setting/details}
    \item[] Question: Does the paper specify all the training and test details (e.g., data splits, hyperparameters, how they were chosen, type of optimizer) necessary to understand the results?
    \item[] Answer: \answerYes{}
    \item[] Justification: The paper specifies the data preprocessing, train/validation/test splits, hyperparameter selection, optimizer settings, and training details needed to understand the results.

\item {\bf Experiment statistical significance}
    \item[] Question: Does the paper report error bars suitably and correctly defined or other appropriate information about the statistical significance of the experiments?
    \item[] Answer: \answerYes{}
    \item[] Justification: The appendix reports mean \(\pm\) sample standard deviation across random seeds for the main experimental sweeps, with the seeds explicitly stated.

\item {\bf Experiments compute resources}
    \item[] Question: For each experiment, does the paper provide sufficient information on the computer resources (type of compute workers, memory, time of execution) needed to reproduce the experiments?
    \item[] Answer: \answerNo{}
    \item[] Justification: The experiments were relatively lightweight and were run on a shared university HPC cluster, but the paper does not systematically report hardware type, memory usage, or wall-clock runtime for each experiment. We do not expect compute requirements to be a limiting factor for reproduction.
    
\item {\bf Code of ethics}
    \item[] Question: Does the research conducted in the paper conform, in every respect, with the NeurIPS Code of Ethics \url{https://neurips.cc/public/EthicsGuidelines}?
    \item[] Answer: \answerYes{}
    \item[] Justification: The research uses a simulated dataset, preserves anonymity, and does not involve human-subject data collection or clinical deployment.

\item {\bf Broader impacts}
    \item[] Question: Does the paper discuss both potential positive societal impacts and negative societal impacts of the work performed?
    \item[] Answer: \answerYes{}
    \item[] Justification: The paper discusses potential positive impacts for dose--response forecasting and glucose-management research, while also noting that real-world clinical deployment could cause harm if inaccurate forecasts were used for treatment decisions without clinical validation, uncertainty calibration, and safety constraints.
    
    
\item {\bf Safeguards}
    \item[] Question: Does the paper describe safeguards that have been put in place for responsible release of data or models that have a high risk for misuse (e.g., pre-trained language models, image generators, or scraped datasets)?
    \item[] Answer: \answerNA{}.
    \item[] Justification: The released artifacts are not high-risk generative models or scraped datasets. The work uses simulated glucose-management data and evaluates forecasting behavior only; it does not release a deployable clinical dosing or decision-support system.

\item {\bf Licenses for existing assets}
    \item[] Question: Are the creators or original owners of assets (e.g., code, data, models), used in the paper, properly credited and are the license and terms of use explicitly mentioned and properly respected?
    \item[] Answer: \answerYes{}
    \item[] Justification: The paper cites the original sources for the existing datasets, simulators, and baseline methods used in the experiments and the lincenses are mentioned.

\item {\bf New assets}
    \item[] Question: Are new assets introduced in the paper well documented and is the documentation provided alongside the assets?
    \item[] Answer: \answerYes{}
    \item[] Justification: The new asset introduced by the paper is the accompanying research code. The supplement documents how to run the proposed method and reproduce the reported experiments.

\item {\bf Crowdsourcing and research with human subjects}
    \item[] Question: For crowdsourcing experiments and research with human subjects, does the paper include the full text of instructions given to participants and screenshots, if applicable, as well as details about compensation (if any)? 
    \item[] Answer: \answerNA{}.
    \item[] Justification: The paper does not involve crowdsourcing experiments or research with human subjects.

\item {\bf Institutional review board (IRB) approvals or equivalent for research with human subjects}
    \item[] Question: Does the paper describe potential risks incurred by study participants, whether such risks were disclosed to the subjects, and whether Institutional Review Board (IRB) approvals (or an equivalent approval/review based on the requirements of your country or institution) were obtained?
    \item[] Answer: \answerNA{}.
    \item[] Justification: The paper does not involve crowdsourcing or research with human subjects; experiments use simulated glucose-management data.

\item {\bf Declaration of LLM usage}
    \item[] Question: Does the paper describe the usage of LLMs if it is an important, original, or non-standard component of the core methods in this research? Note that if the LLM is used only for writing, editing, or formatting purposes and does \emph{not} impact the core methodology, scientific rigor, or originality of the research, declaration is not required.
    \item[] Answer:  \answerNA{}.
    \item[] Justification: LLMs are not used as an important, original, or non-standard component of the core method. Any LLM use, if applicable, was limited to writing, editing, or formatting support.

\end{enumerate}
\end{document}